\pdfoutput=1

\documentclass[11pt]{article}

\usepackage[final]{acl}
\usepackage{times}
\usepackage{latexsym}
\usepackage{booktabs}
\usepackage{multirow}
\usepackage{xcolor}
\usepackage{amsmath}
\usepackage{amssymb}
\usepackage[T1]{fontenc}

\usepackage[utf8]{inputenc}

\usepackage{microtype}

\usepackage{inconsolata}

\usepackage{graphicx}
\usepackage{multirow}
\usepackage{booktabs}
\usepackage{graphicx}
\usepackage{subcaption}
\title{CAI: Caption-Sensitive Attention Intervention for Mitigating Object Hallucination in Large Vision-Language Models}

\author{Qiming Li$^{1}$\thanks{~Equal Contribution}, Zekai Ye$^{1}$\footnotemark[1], Xiaocheng Feng$^{1,2}$, Weihong Zhong$^1$, Libo Qin$^3$, Ruihan Chen$^1$, \\
\textbf{Baohang Li$^1$, Lei Huang$^1$, Baohang Li$^1$, Kui Jiang$^1$, Yaowei Wang$^2$, Ting Liu$^1$, Bing Qin$^{1,2}$}\\
  $^{1}$Harbin Institute of Technology\quad \quad \quad $^2$Peng Cheng Laboratory
  \quad \quad \quad$^{3}$Central South University\\
  \texttt{\{qmli,zkye\}@ir.hit.edu.cn}
  \\
}

\begin{document}
\maketitle
\begin{abstract}
Although Large Vision-Language Models (LVLMs) have demonstrated powerful capabilities in interpreting visual information, they frequently produce content that deviates from visual information, leading to object hallucination. 
To tackle this, recent works mostly depend on expensive manual annotations and training cost, or significantly increase inference time.
In this work, we observe that LVLMs' attention to visual information is significantly stronger when answering caption queries compared to non-caption queries.
Inspired by this phenomenon, we propose \textbf{C}aption-sensitive \textbf{A}ttention \textbf{I}ntervention (\textbf{CAI}), a training-free, plug-and-play hallucination mitigation method that leverages the attention activation pattern in response to caption queries to enhance LVLMs' visual perception capability. 
Extensive experimental results across four benchmarks covering both discriminative and generative tasks, demonstrate that CAI achieves state-of-the-art (SOTA) hallucination mitigating performance only with minimal additional inference cost.

\end{abstract}

\section{Introduction}

Despite the continuous advancements in the performance of large vision-language models (LVLMs) in recent years, it is widely observed that LVLMs frequently generate content that conflicts with the corresponding visual information, leading to hallucination \cite{sahoo2024comprehensive,huang2023survey}.

\begin{figure}[t]
  \includegraphics[width=\columnwidth]{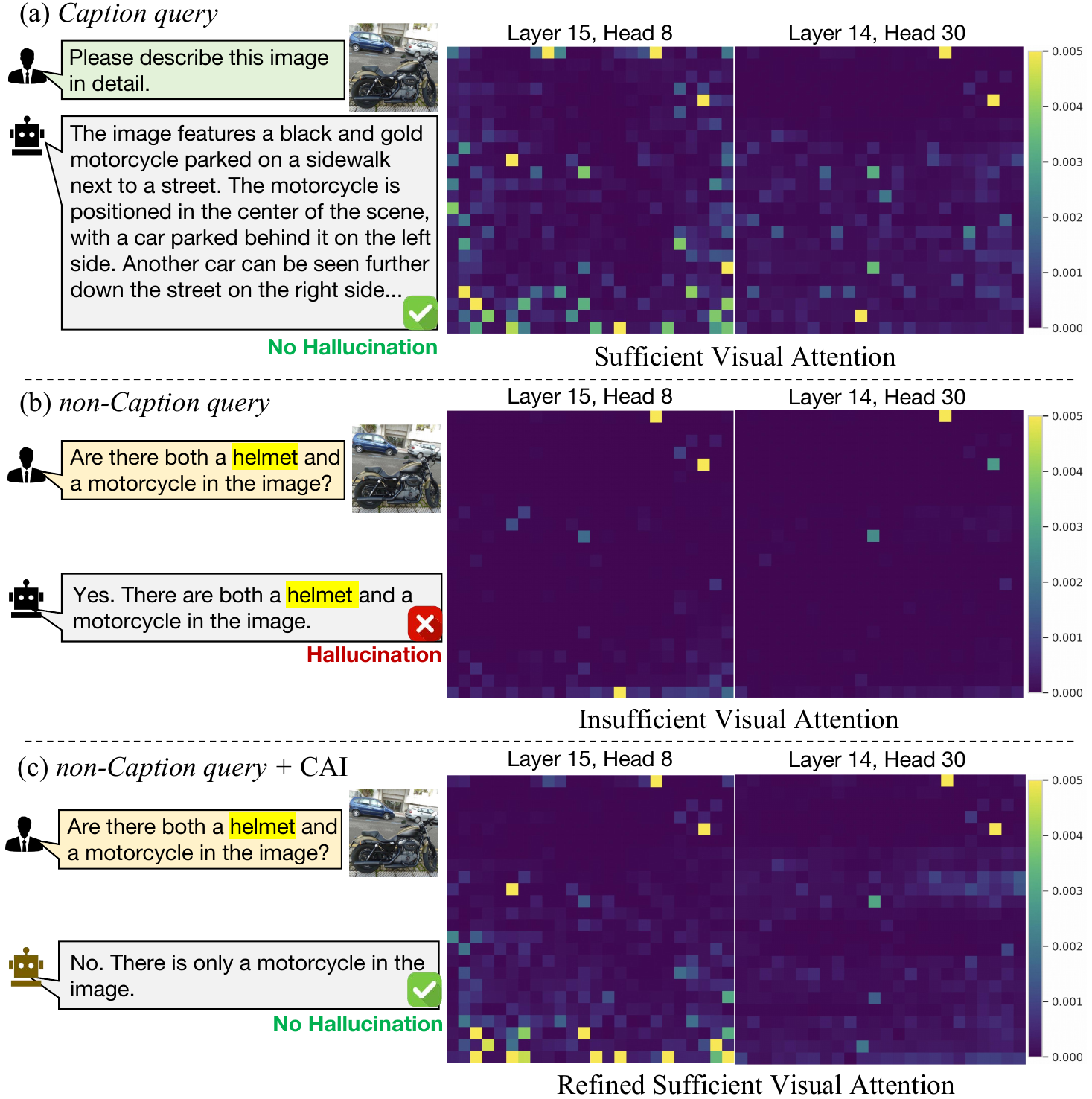}
  \caption{The visualization of attention weights at image patch level across different conversations. LVLM correctly generates the detailed content of the image in response to the caption query, but exhibits hallucination (e.g., "helmet") when answering the non-caption query. CAI refines LVLM's visual attention patterns from insufficient to sufficient, effectively enhancing visual perception capability and mitigating hallucination.
  }
  \label{fig:figure1}
\end{figure}

Previous works \cite{arif2025fixing,Bi2024UnveilingVP} show that LVLMs' insufficient attention to visual information is considered a key factor in hallucination.
To tackle this, recent works for mitigating hallucination mostly use contrastive decoding strategy \cite{leng2024mitigating,zhong2024investigatingmitigatingmultimodalhallucination} which arises high inference latencies, or train LVLMs using carefully designed data \cite{you2023ferret,yu2024rlhf} which incurs expensive manual annotation and computation cost.

To address the aforementioned limitations, 
we focus on exploring how to enhance LVLMs' perception capability by providing sufficient attention to visual information.
In this work, as shown in Figure 1 (a) and (b), we reveal a critical phenomenon: visual attention across particular attention heads 
was significantly enhanced when fed caption compared to non-caption queries. We term these attention heads as caption-sensitive attention heads. As an enhancement of their visual attention leads to a corresponding improvement in LVLM's perception capability, we believe that these heads are also visually sensitive. Therefore, we manage to probe and refine these caption-sensitive attention heads. 

Inspired by the aforementioned phenomenon,
we propose \textbf{C}aption-sensitive \textbf{A}ttention \textbf{I}ntervention (\textbf{CAI}), a training-free, plug-and-play method, to refine caption-sensitive attention heads outputs during inference to enhance LVLMs' fine-grained visual perception capability and mitigate hallucination.
First, we identify the optimal caption query from candidates, which activates the model's inherent visual perception capability with the minimal necessary attention weight shift cost.
Secondly, following previous work \cite{li2024inference}, we train binary classifiers to identify caption-sensitive attention heads and compute their attention output heads shifts, which quantify the differences from non-caption to caption queries and serve as a vision-centric optimization direction.
Finally, we apply the precomputed attention output shifts to intervene caption-sensitive attention heads during inference. 
As shown in Figure \ref{fig:figure1} (b) and (c), after using CAI, LVLM enhances sufficient visual attention and effectively mitigates hallucination.

We evaluate the performance of CAI across multiple discriminative and generative tasks, using models such as LLaVA-1.5-7b \cite{liu2024improved}, Qwen-VL-Chat \cite{bai2023qwen}, and LLaVA-NeXT \cite{liu2024llavanext}. On the POPE \cite{li2023evaluating} benchmark, the accuracy and the F1 score improve by 5.14\% and 5.50\% on average. On the MME \cite{Fu2023MMEAC} hallucination subset, the scores increase by 64.3 points on average. Furthermore, hallucination rates decrease by 7.8\% on the MMHalBench \cite{sun2023aligning}, while the informativeness of the generated responses improves. 

The main contributions can be summarized as: 

(1) Our work is the first to explicitly reveal the impact of caption versus non-caption queries on the attention activation patterns of LVLMs.

(2) We propose \textbf{CAI}, a training-free, plug-and-play method significantly mitigates object hallucination in LVLMs by refining caption-sensitive attention head outputs during the inference.


(3) Comprehensive experimental results demonstrate that CAI effectively mitigates hallucination.

\section{Quantitative Analysis of the Effect of Caption Queries on Visual Attention}

To better validate the motivation of our CAI method that caption queries can help LVLMs refine visual attention activation patterns, we constructed a quantitative analysis experiment. We sample 1,000 images from the MS-COCO dataset \cite{lin2014microsoft}. For each image, we propose one caption query and two different non-caption queries (non-caption-1 \& non-caption-2) to analyze differences attributable to query types. We compute the $Change\ Rate$ to quantify differences in visual attention weight changes.
The specific calculation and other experimental details can be found in the Appendix \ref{sec:G}.

\begin{figure}[!ht]
  \includegraphics[width=\columnwidth]{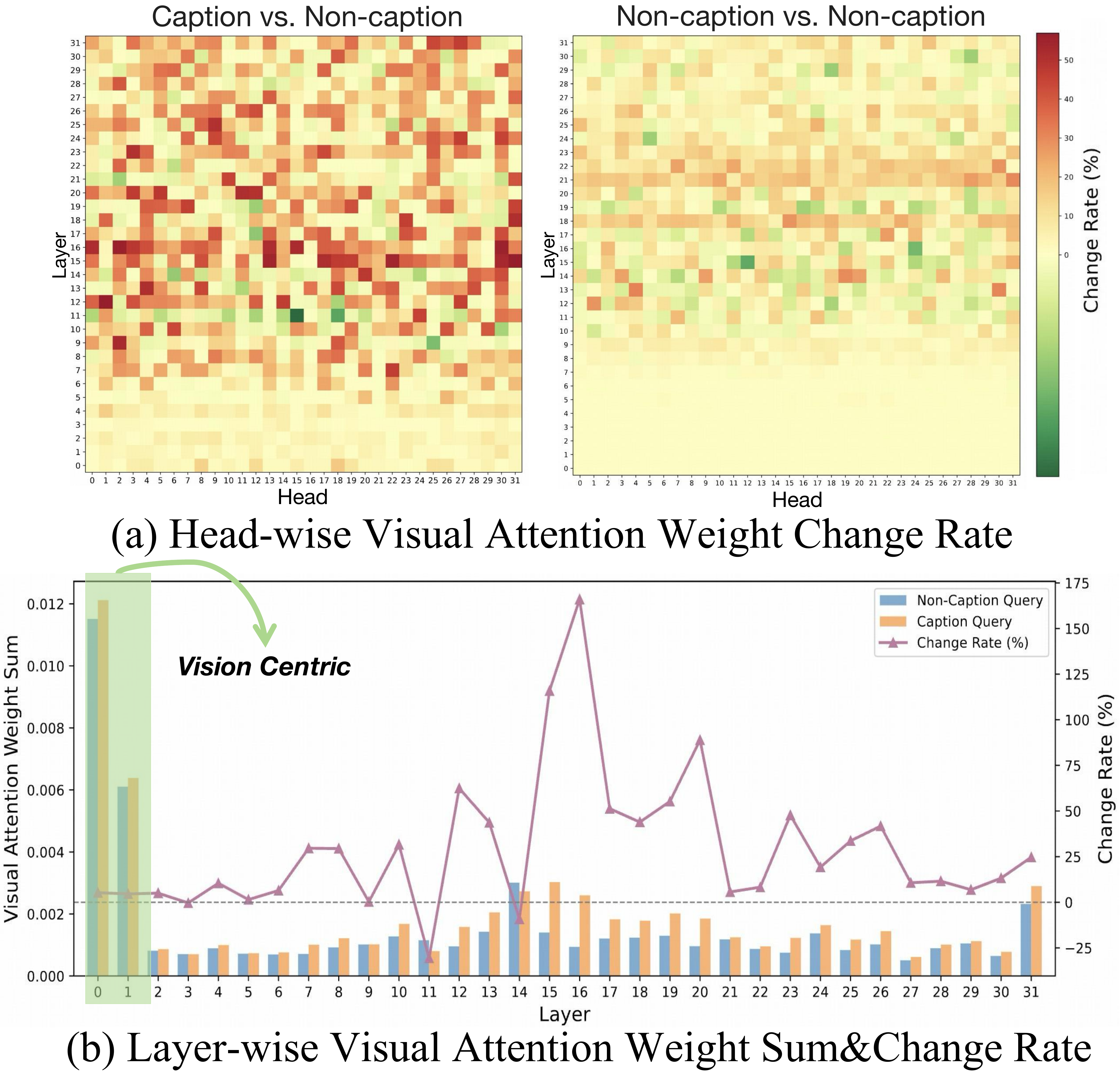}
  \caption{A systematic quantitative analysis from head-wise (a) and layer-wise (b) on visual attention weights. The comparison shows that the caption query significantly enhanced the visual attention of LLaVA-1.5-7b. 
  }
  \label{fig:quan}
\end{figure}

As illustrated in Figure \ref{fig:quan}, caption queries have a more significant impact on the LVLM's visual attention weights compared with non-caption queries. 65.92\% of the attention heads and 30 of the 32 layers exhibit a enhancement in visual attention weight. As shown in Figure \ref{fig:quan} (b), the first two layers allocate significantly more visual attention compared to other layers, suggesting that they are centered on visual information and relatively insensitive to input queries. In deeper layers, LVLM's enhanced visual attention is an important and non-negligible reason for stronger perception capability.

\section{Methods}

\subsection{Task Formulation}
 
We consider a LVLM parametrized by \( \theta \). The model receives as input a visual input $ \boldsymbol{V} = \{v_1, v_2, \dots, v_m\} $ and a textual query $ \boldsymbol{T} = \{t_1, t_2, \dots, t_n\} $, where $ m $ and $ n $ denote the sequence lengths of the visual input and textual inputs. The textual and visual inputs are concatenated together to form the first layer input $ \boldsymbol{H}^{1} = \mathrm{concat}(\boldsymbol{V},\boldsymbol{T}) \in \mathbb{R}^{(m+n)\times d}$ for the $L$ layers $\times$ $H$ heads language decoder. 

\begin{figure*}[!ht]
  \includegraphics[width=1.0\linewidth]{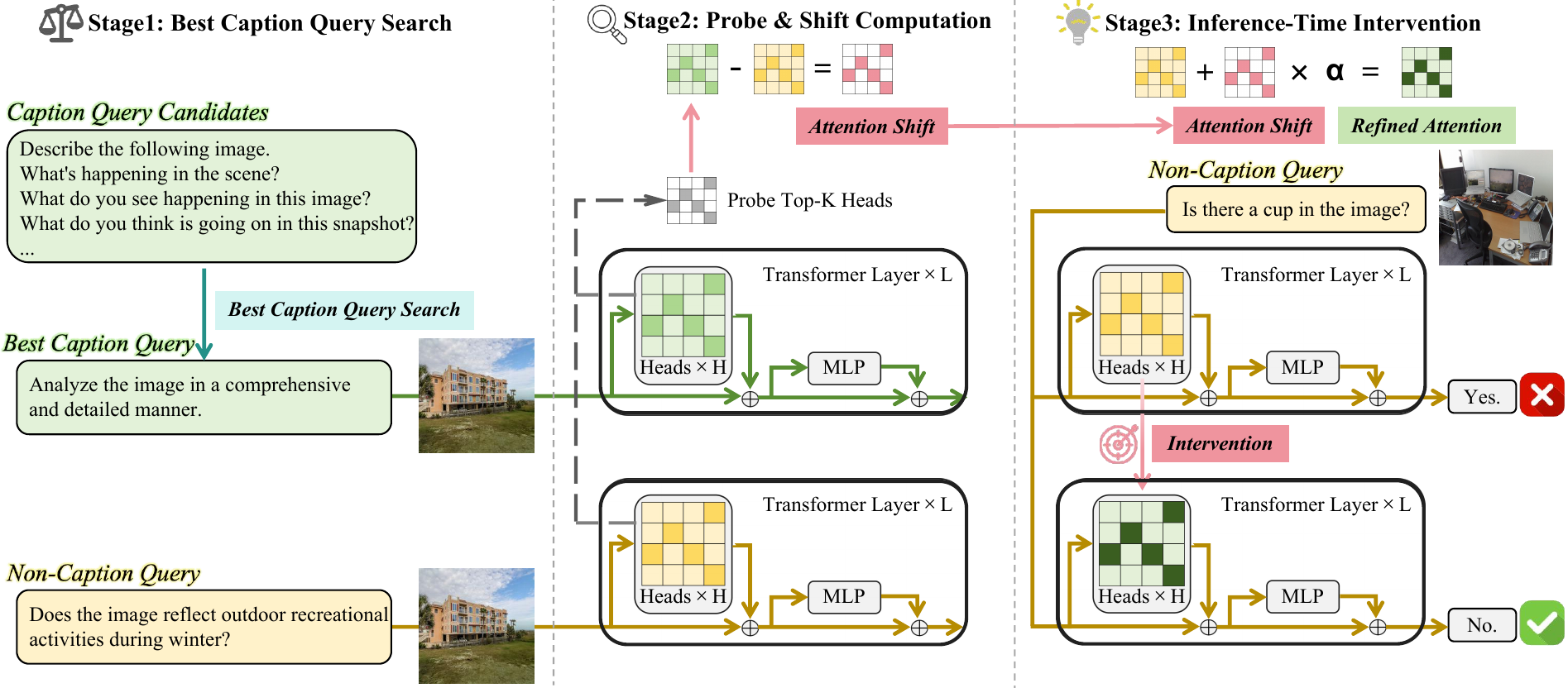}
  \caption{An overview of the CAI method. Each square in the matrix represents the attention head output. Squares with dark green color indicate refined attention head outputs. CAI consists of three stages: (1) \S\ref{search} Best caption query search algorithm is designed to seek the best optimization target query with minimal necessary attention weight shift. (2) \S\ref{probe}  The original and modified attention outputs are used to identify caption-sensitive attention heads and compute attention output shift vectors. (3) \S\ref{intervention} Precomputed attention shift vectors are applied to the top $K$ caption-sensitive attention heads during inference, thereby enhancing their visual attention and activating the model's inherent fine-grained visual perception to mitigate hallucination. }
  \label{fig:method}
\end{figure*}
During the forward pass, the input $ \boldsymbol{H}^{l}$ received by the $h$-th attention head at $l$-th layer is linearly transformed using independent weight matrices to generate the Query, Key and Value matrices, denoted as $\boldsymbol{Q}_{(l,h)}\in\mathbb{R}^{(m+n)\times d}$, $\boldsymbol{K}_{(l,h)}\in\mathbb{R}^{(m+n)\times d}$ and
$\boldsymbol{V}_{(l,h)}\in\mathbb{R}^{(m+n)\times d}$,

where $d$ denotes the head hidden dimensions. The generated Query, Key, and Value matrices are then used to compute the attention score, attention weight matrix and attention output as follows: 
\begin{equation}
\boldsymbol{\dot{A}}_{(l,h)} = \frac{\boldsymbol{Q}_{(l,h)} \boldsymbol{K}_{(l,h)}^T}{\sqrt{d}},
\boldsymbol{A}_{(l,h)} = \textrm{softmax}(\boldsymbol{\dot{A}}_{(l,h)}),
\notag
\label{eq:attention_relevance}
\end{equation}
\begin{equation}
\boldsymbol{O}_{(l,h)} = \boldsymbol{A}_{(l,h)}\boldsymbol{V}_{(l,h)}.
\label{eq:attention_output}
\end{equation}
At each layer, the updated hidden state $\boldsymbol{H}^{l+1}$ is then computed by adding the residual connection to the output of the multi-head attention mechanism:
\begin{equation}
    \boldsymbol{H}^{l+1} = \boldsymbol{H}^{l} + \sum_{h=1}^{H} \boldsymbol{O}_{(l,h)} \cdot \boldsymbol{W}_o^l,
\end{equation}
where $\boldsymbol{W}_o^l\in\mathbb{R}^{Hd\times d}$ is the learnable weight matrix for the linear transformation applied after concatenating the outputs from all $H$ attention heads.
Finally, the model predicts the next token in an autoregressive manner based on the last layer output.

\subsection{Best Caption Query Search Algorithm}
\label{search}

This module aims to seek the best caption query, which induces the minimal necessary attention weight shift to activate the LVLMs' fine-grained visual perception capabilities. For a single VQA question, we separately use a certain caption query $T$ from $J$ candidate queries and a non-caption query $T'$ paired with same image $V$ as inputs during the forward pass to compute caption-sensitive attention weight matrix $\boldsymbol{A}_{(l,h)}$ and non-caption attention weight matrix $\boldsymbol{A'}_{(l,h)}$. The attention weight shift matrix can be computed as:
\begin{equation}
\boldsymbol{A}_{shift}=\sum_{l=1}^{L}\sum_{h=1}^{H}\left({\boldsymbol{A}_{(l,h)}-\boldsymbol{A'}_{(l,h)}}\right).
\end{equation}
For a VQA dataset with a batch size of $B$ and a caption query candidate list with length of $J$, the index of the best query $j$ in the candidate list can then be calculated as:
\begin{equation}
\mathop{\arg\min}\limits_{j}\ \sum_{b=1}^{B}{\boldsymbol{A}^{b,j}_{shift}}\ \ \  s.t.\ j\in J,
\end{equation}
where $\boldsymbol{A}^{b,j}_{shift}$ denotes the attention weight shift matrix when answering the $b$-th VQA using the j-th caption query.

\subsection{Caption-Sensitive Attention Heads Probe}
\label{probe}

This module aims to identify caption-sensitive attention heads, which exhibit significant differences in attention outputs when responding to caption and non-caption queries.
We focus on the LVLMs' attention output shift of visual information, aiming to minimize the influence of textual semantic information during the probing process. To achieve this, we set the last token's text-related attention scores of each attention head to $-\infty$ during the forward pass, and compute the modified attention output:
\begin{equation}
    \boldsymbol{\dot{A}}_{(l,h)}[m:m+n] = -\infty, 
\end{equation}
\begin{equation}
   {\boldsymbol{\hat{O}}_{(l,h)} }=  \textrm{softmax}({\boldsymbol{\dot{A}}_{(l,h)}})\boldsymbol{V}_{(l,h)},
\end{equation}
\begin{equation}
   {\boldsymbol{\widetilde{O}}_{(l,h)} =  {\boldsymbol{\hat{O}}_{(l,h)} }[m+n]}.
\end{equation}

For a dataset with a batchsize of $B$, the last token's modified attention output of $b'$-th VQA problem when answering the caption query and non-caption query are denoted as $\boldsymbol{\widetilde{O}}^{b}_{(l,h)}$ and $\boldsymbol{\widetilde{O'}}^{b}_{(l,h)}$. Respectively, the last token's origin attention output are denoted as $\boldsymbol{{O}}^{b}_{(l,h)}$ and $\boldsymbol{{O'}}^{b}_{(l,h)}$. 

For each head, we use the $B$ pairs of modified attention output
as input to train a binary classifier $f_{l,h}(\cdot)$ that predicts wether the input sentence is a caption query:
\begin{equation}
\mathop{\arg\min}\limits_{{f_{l,h(\cdot)}}}\sum_{b=1}^{B} \mathcal{L}\left(f_{l,h}\left(x_b\right), y_b\right),
\end{equation}
where $x_b\in\{\boldsymbol{\widetilde{O}}^{b}_{(l,h)} ,\boldsymbol{\widetilde{O'}}^{b}_{(l,h)}\}$ denotes the input of classifier and $y_b\in\{0,1\}$ denotes the category of query.
We then select the top $K$ binary classifiers with the highest accuracy. In order to compute the optimization direction for each head, the attention output shift vector is computed as follows:
\begin{equation}
   \boldsymbol{S}_{(l,h)} = \frac{1}{B} \sum_{b=1}^{B}\left( \boldsymbol{O}_{(l,h)}^b - \boldsymbol{O'}_{(l,h)}^{b}\right).
\end{equation}

\subsection{Intervention at Inference Time}
\label{intervention}
This module aims to refine the top $K$ attention heads output that are most sensitive to caption queries at inference-time. We leverage the precomputed shift vectors to refine these heads from insufficient visual attention states to sufficient states, thereby enhancing the model’s fine-grained visual perception capability and mitigate hallucination. At each layer, the updated hidden state after intervention is computed as:
\begin{equation}
    \boldsymbol{H}^{l+1} = \boldsymbol{H}^{l} + \sum_{h=1}^{H} \left(\boldsymbol O_{(l,h)}+\mathbb{I}_{(l,h)}\alpha\boldsymbol{S}_{(l,h)}\right) \cdot \boldsymbol{W}_o^l,
\end{equation}
where $\mathbb{I}_{(l,h)}$ is a gating function, assigning a value of $1$ to attention heads with top $k$ highest accuracy, and 0 to the others. $\alpha$ represents the intensity of the intervention.

\section{Experiments}

\subsection{Experimental Setup}

We comprehensively evaluate the methods for both discriminative and generative tasks to measure the effectiveness and robustness of the methodes.

\noindent\textit{\textbf{Discriminative Tasks:}}

\noindent\textbf{POPE} employs a binary question-answering format, inquiring LVLMs to answer if a special object exists in the given image.
Following previous works, we adopt Accuracy and F1 score as the evaluation metrics.

\noindent\textbf{MME} serves as a comprehensive tool for assessing the capabilities of LVLMs across both 10 perception tasks and 4 cognition tasks. 
Consequently, task scores are reported as the evaluation metrics.

\begin{table*}[!ht]
\small
\renewcommand{\arraystretch}{0.95}
\centering
\setlength{\tabcolsep}{5pt}
\begin{tabular}{clccccccc}
\toprule
\multirow{2}{*}{\textbf{Setting}} & \multirow{2}{*}{\textbf{Method}} & \multicolumn{2}{c}{\textbf{LLaVA-1.5-7b}} & \multicolumn{2}{c}{\textbf{Qwen-VL-Chat}} & \multicolumn{2}{c}{\textbf{LLaVA-NeXT}}\\
\cmidrule(lr){3-4} \cmidrule(l){5-6} \cmidrule(l){7-8}
 & & Accuracy & F1 Score & Accuracy & F1 Score &Accuracy & F1 Score\\
\midrule
\multirow{6}{*}{Random} 
& Regular & 83.29 & 81.33 & 84.63 & 82.61 & 84.78 & 86.43 \\
& VCD & 87.73 & 87.16 & 86.93 & 85.46 & 88.76 & 89.57 \\
& OPERA & 89.20 & 88.81 & 85.71 & 84.64 & 90.27 & 89.71 \\
& PAI & 86.33 & 84.56 & 85.38 & 85.54 & 88.40 & 87.16 \\
& VTI & 89.50 & 88.89 & 86.73 & 85.59 & 89.23 & 88.68 \\
& CAI(ours) & \textbf{89.87}\textcolor{red}{(+6.58)} & \textbf{89.43}\textcolor{red}{(+8.10)} & \textbf{88.17}\textcolor{red}{(+3.54)} & \textbf{87.31}\textcolor{red}{(+4.70)} & \textbf{90.68}\textcolor{red}{(+5.90)} & \textbf{90.42}\textcolor{red}{(+3.99)} \\
\cmidrule(lr){2-8}
\multirow{6}{*}{Popular}
& Regular & 81.88 & 80.06 & 83.63 & 81.53 & 83.23 & 84.77 \\
& VCD & 85.38 & 85.06 & 85.17 & 83.68 & 87.01 & 87.70 \\
& OPERA & 86.64 & 86.62 & 84.82 & 83.99 & 87.16 & 87.68 \\
& PAI & 85.33 & 83.62 & 84.20 & 83.10 & 86.65 & 86.99 \\
& VTI & 87.36 & 86.69 & 85.67 & 84.48 & 87.33 & 87.16 \\
& CAI(ours) & \textbf{88.32}\textcolor{red}{(+6.44)} & \textbf{87.95}\textcolor{red}{(+7.89)} & \textbf{87.73}\textcolor{red}{(+4.10)} & \textbf{86.84}\textcolor{red}{(+5.31)} & \textbf{89.53}\textcolor{red}{(+6.30)} & \textbf{89.24}\textcolor{red}{(+4.47)} \\
\cmidrule(lr){2-8}
\multirow{6}{*}{Adversarial}
& Regular & 78.96 & 77.57 & 81.03 & 79.30 & 81.19 & 82.50 \\
& VCD & 80.88 & 81.33 & 83.10 & 82.04 & 84.80 & 85.23 \\
& OPERA & 81.24 & 81.38 & 82.67 & 79.89 & 85.20 & 85.54 \\
& PAI & 83.17 & 81.67 & 82.19 & 82.06 & 84.32 & 83.68 \\
& VTI & 82.57 & 82.11 & 83.13 & 82.16 & 85.35 & 84.52 \\
& CAI(ours) & \textbf{84.27}\textcolor{red}{(+5.31)} & \textbf{84.41}\textcolor{red}{(+6.84)} & \textbf{84.33}\textcolor{red}{(+3.30)} & \textbf{83.92}\textcolor{red}{(+4.62)} & \textbf{85.97}\textcolor{red}{(+4.78)} & \textbf{86.07}\textcolor{red}{(+3.57)} \\
\bottomrule
\end{tabular}
\caption{Main results on POPE tasks. We evaluate the accuracy and F1 Score of various LVLMs on the MS-COCO POPE tasks. The best performances within each setting are bolded. \textbf{CAI(ours)} demonstrates the best hallucination mitigation performance among several methods.}
\label{tab:main_result}
\end{table*}

\begin{table*}[!ht] 
\small
\renewcommand{\arraystretch}{0.90} 
\setlength{\tabcolsep}{3pt} 
\centering
\begin{tabular}{lccccccccccccccc}
\toprule
\multirow{2}{*}{\textbf{Method}}  & \multicolumn{5}{c}{\textbf{LLaVA-1.5-7b}} & \multicolumn{5}{c}{\textbf{Qwen-VL-Chat}} & \multicolumn{5}{c}{\textbf{LLaVA-NeXT}} \\
\cmidrule(lr){2-6} \cmidrule(l){7-11} \cmidrule(l){12-16}
& Exist. & Count & Pos. & Color &Total& Exist. & Count & Pos. & Color & Total & Exist. & Count & Pos. & Color & Total \\
\midrule 
Regular &175.7& 124.7  & 114.0 & 151.0&565.4 & 170.0  & 135.0 & 123.3 & 170.0& 598.3 &180.0 &105.0 &150.0 & 151.7 & 586.7  \\
VCD&180.3&131.7  & 125.0 & 155.0 & 592.0   & 180.0&133.3&131.7&175.0& 620.0 & 185.0 &125.0 & 133.3 & 168.3 &611.6  \\
OPERA&165.0& 116.0 & 133.3 & 149.0 & 563.3  &  180.0 &140.0&138.3&175.0&633.3 & 183.8 & 121.3 & 155.0 &162.1& 622.2 \\
PAI &190.0 & \textbf{148.3} &126.7 & 160.0 & 625.0 & 175.0 & 141.6 & 132.5 & 177.5 & 626.6 & 185.0 & 128.3 & 148.3 & 170.8 & 632.4 \\
VTI & 185.0 & 140.0 &135.0 &165.7 & 619.0 & 180.0 & 142.5 & 133.0 & 178.0 & 633.5 & 186.7 & 126.7 & 150.0 & 172.5 & 635.9 \\
CAI(ours)&\textbf{190.0}&141.6  &\textbf{140.0}  &\textbf{170.0}  & \textbf{641.6}  & \textbf{185.0} &\textbf{150.0}&\textbf{133.3}&\textbf{180.0}&\textbf{648.3} & \textbf{190.0} & \textbf{133.3}& \textbf{155.0 }&\textbf{175.0}& \textbf{653.3} \\
\bottomrule
\end{tabular}
\caption{Main results on the hallucination subset of MME. The best performances within each setting are bolded.  }
\label{tab:mme_result}
\end{table*}

\noindent\textit{\textbf{Generative Tasks:}}

\noindent\textbf{CHAIR} \cite{rohrbach2018object} is a widely used metric for assessing object hallucination in responses of LVLMs. Following previous work, We use the MS-COCO Chair subset with the prompt "Please describe this image in detail." to evaluate the hallucination mitigating capabilities of LVLMs. The CHAIR metric comprises two important indicators, denoted as $\textrm{CHAIR}_i$ and $\textrm{CHAIR}_s$, with the following calculation formulas:
\begin{equation}
\small
   \textrm{CHAIR}_i = \frac{|\{\textrm{Hallucinationted}\ \textrm{objects}\}|}{|\{\textrm{All}\ \textrm{objects}\ \textrm{mentioned}\}|} \notag
\end{equation}
\begin{equation}
\small
   \textrm{CHAIR}_s = \frac{|\{\textrm{Sentence with hallucination objects}\}|}{|\{\textrm{All sentence}\}|} \notag
\end{equation}

\noindent\textbf{MMHal-Bench} comprises 96 meticulously designed VQA questions, which evaluates response-level hallucination rate and informativeness. It asks GPT-4 to compare model outputs with human responses and object labels for evaluation.

\noindent\textbf{Baselines.} We adopt the widely used LLaVA-1.5-7b, Qwen-VL-Chat and LLaVA-NeXT \cite{liu2024llavanext} as our baseline LVLMs. We compared CAI with the following SOTA training-free methods:

\textbf{(1) Baselines tailored for decoding:} 
VCD \cite{leng2024mitigating} contrasts model logits derived from original and distorted visual input to reduce the over-reliance on statistical bias and unimodal priors. OPERA \cite{huang2024opera} introduces a penalty term on the model logits.

\textbf{(2) Baselines utilizing inference-time intervention:} 
PAI \cite{liu2024paying} intervenes on attention heads by leveraging their original direction and optimizes the output distribution during decoding to mitigate language bias. VTI \cite{liu2024reducing} mitigates hallucination by steering layer-level latent space representations during inference to enhance visual feature stability. 

Despite prior findings \cite{bi2024unveiling} indicating the significant role of attention heads in visual perception, there is a lack of approaches that analyze at head level and do not rely on specific decoding strategies (which increase inference time). To address these limitations, our CAI probes and refines attention heads with minimal additional inference cost by exploiting the differential attention activation pattern between caption and non-caption queries, thereby achieving superior results.

\noindent\textbf{Implementation Details.} In our experiments, we utilized 13 caption queries and 100 VQA from LLaVA pretrain dataset to search the best caption. Then we utilized 1000 VQA from LLaVA pretraining dataset pairs with the searched best caption query and non-caption queries to identify caption-sensitive attention heads and computed the attention shift vectors. For each attention head, SVM \cite{cortes1995support} was used as the classifier and 2-fold cross-validation was performed to evaluate its accuracy. 
More detailed experimental procedures are provided in Appendix \ref{sec:A}.

\subsection{Main Results}
Comprehensive evaluations demonstrate that our method exhibits superior hallucination mitigation capabilities in discriminative and generative tasks.

\noindent\textbf{Result on POPE.} Table \ref{tab:main_result} presents the POPE tasks results under three different experimental settings. 

\begin{figure}[!ht]
  \includegraphics[width=\columnwidth]{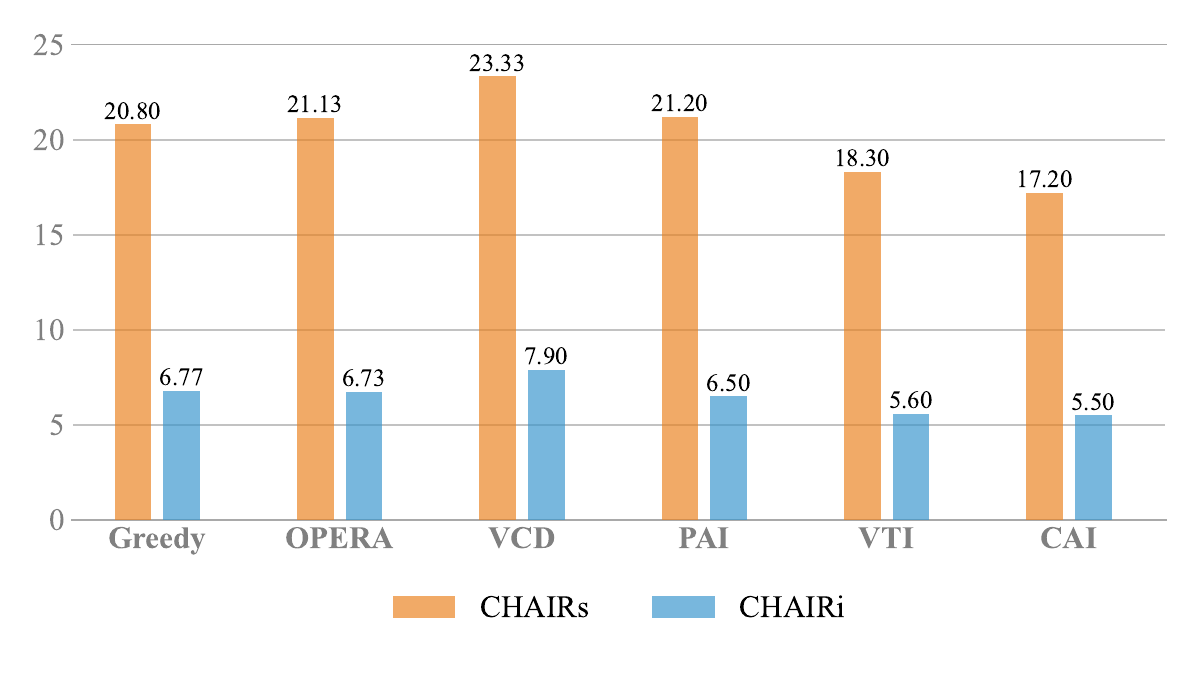}
  \caption{Main result of LLaVA-1.5-7b on MS-COCO CHAIR task. Smaller values of $\mathrm{CHAIR}_i$ and $\mathrm{CHAIR}_s$ indicate that the method demonstrates stronger hallucination mitigation capabilities at instance and sentence levels. ${Max\_new\_tokens}$ is set to be 64.}
  \label{fig:chair}
\end{figure}

\noindent\textbf{(1) SOTA performence.} Compared with other methods, CAI achieved superior hallucination mitigation performance across all experimental configurations. 
Specifically, it leads to an average improvement of 6.11\% in accuracy and 7.61\% in F1 score for LLaVA-1.5-7b, 3.65\% and 4.88\% for Qwen-VL-Chat, 5.66\% and 4.01\% for LLaVA-NeXT, resulting in SOTA hallucination mitigation effects. 
\textbf{(2) Easy to deploy in open-source LVLMs.} 
During the best caption query searching, caption-sensitive attention heads probe and shift vector computation stages, 
CAI selected images, caption query candidates and non-caption queries from LLaVA-1.5-7b pre-training dataset, which are outside the domain of the benchmark datasets. Despite this cross-domain discrepancy, our method achieved significant improvements even when applied to Qwen-VL-Chat and LLaVA-NeXT, indicating that CAI does not rely on specific models or data and deploy in open-source LVLMs easily.

\noindent\textbf{Results on MME.}
Table \ref{tab:mme_result} and Figure \ref{fig:mme_full} respectively present the experimental results for the MME hallucination subset and full set. \textbf{Our method effectively mitigates hallucination while preserving the LVLMs' other foundational capabilities.} On the MME hallucination subset, our method achieved the best improvements across all capabilities with score increases of 76.2 for LLaVA-1.5-7b, 50.0 for Qwen-VL-Chat and 66.6 for LLaVA-NeXT. On the full MME dataset, performance improved on 13 out of 14 perception and reasoning tasks, with an overall score increase of 197.63 for LLaVA-1.5-7b.

\begin{table}[!ht] 
\small
\centering
\renewcommand{\arraystretch}{1.0} 
\setlength{\tabcolsep}{2pt} 

\begin{tabular}{lcccc}
\toprule
\multirow{2}{*}{ \textbf{Method}} & \multicolumn{2}{c}{\textbf{LLaVA-1.5-7b}} & \multicolumn{2}{c}{\textbf{Qwen-VL-Chat}} \\
\cmidrule(lr){2-3}
\cmidrule(lr){4-5}
& Score\textbf{$\uparrow$} & VH Rate\%\textbf{$\downarrow$} & Score\textbf{$\uparrow$} & VH Rate\%\textbf{$\downarrow$} \\
\midrule
Greedy & 1.86 & 63.5 & 2.93 & 41.1 \\
VCD & 2.12 & 54.2 &  2.77 & 39.2 \\
OPERA & 2.15 & 54.2 & 2.94 & 38.4\\
PAI & 2.27 & 53.2 &  2.87 & 39.5 \\
VTI & 2.33 & 52.2 & 2.99 & 38.4\\

CAI(ours) & 2.43 & 51.0 &  3.04 & 38.0 \\
\bottomrule
\end{tabular}
\caption{Main result on MMHal-Bench. 
}
\label{mmhbench-result}
\end{table}

\noindent\textbf{Results on CHAIR.}
Figure \ref{fig:chair} demonstrates that our method significantly reduces both sentence-level and instance-level hallucination in responses to caption queries. Specifically, we observed reductions of 3.6\% in the $\textrm{CHAIR}_s$ metric and 1.27\% in the $\textrm{CHAIR}_i$ metric. 
By employing the CAI, we precisely identify the attention heads that play a crucial role in visual perception under the caption task and accurately estimate their optimization directions. 
Although these attention heads have already been activated and demonstrate the ability to perceive visual information, applying CAI method can further strengthen visual attention, enhancing LVLMs' perceptual capabilities and resulting in better performance on the caption task.

\noindent\textbf{Results on MMHal-Bench.}
Table \ref{mmhbench-result} presents our method effectively reduces the hallucination rate in responses to non-caption queries while enhancing informativeness, outperforming several inference-time intervention methodes.

\section{Analysis and Discussions}
\subsection{Inference Latency}
\begin{table}[htbp] 
\small
\centering
\renewcommand{\arraystretch}{1.0} 
\setlength{\tabcolsep}{6pt} 

\begin{tabular}{lrrc}
\toprule
Method & TTFT(ms) & TPOT(ms) & Acc(\%)\\
\midrule
LLaVA-1.5-7b & 99.8 \textcolor{gray}{{1.0}\(\times\)} & 36.0 \textcolor{gray}{1.0\(\times\)} & 78.96 \\
\midrule
+VCD & 160.1 \textcolor{red}{1.6\(\times\)} & 96.8 \textcolor{red}{2.7\(\times\)} & 80.88 \\
+OPERA & 109.8 \textcolor{red}{1.1\(\times\)} & 69.5 \textcolor{red}{1.9\(\times\)} & 81.24\\
+VDGD & 377.8 \textcolor{red}{3.8\(\times\)} & 340.9 \textcolor{red}{9.5\(\times\)} & 65.82 \\
\midrule
+CAI(ours) & 102.2 \textcolor[rgb]{0,0.5,0}{1.0\(\times\)} & 36.5 \textcolor[rgb]{0,0.5,0}{1.0\(\times\)}  & 84.50 \\
\bottomrule
\end{tabular}
\caption{Inference latency (Time to First Token, Time Per Output Token) and the accuracy on MS-COCO adversarial POPE of different methods.
}
\label{lantency}
\end{table}
As shown in Table \ref{lantency}, although VDGD \cite{ghosh2024vdgdmitigatinglvlmhallucinations} attempts to mitigate hallucination using captioning capability, directly using caption description significantly increases computational cost and forces the model to process longer context, leading to a performance drop on the POPE task. Compared to contrastive decoding-based methods which trade-off speed for accuracy, CAI implicitly utilizes captioning capabilities, achieveing faster inference speed and better performance.
\subsection{Necessity of the Search Algorithm}

\begin{figure}[!ht]
  \includegraphics[width=\columnwidth]{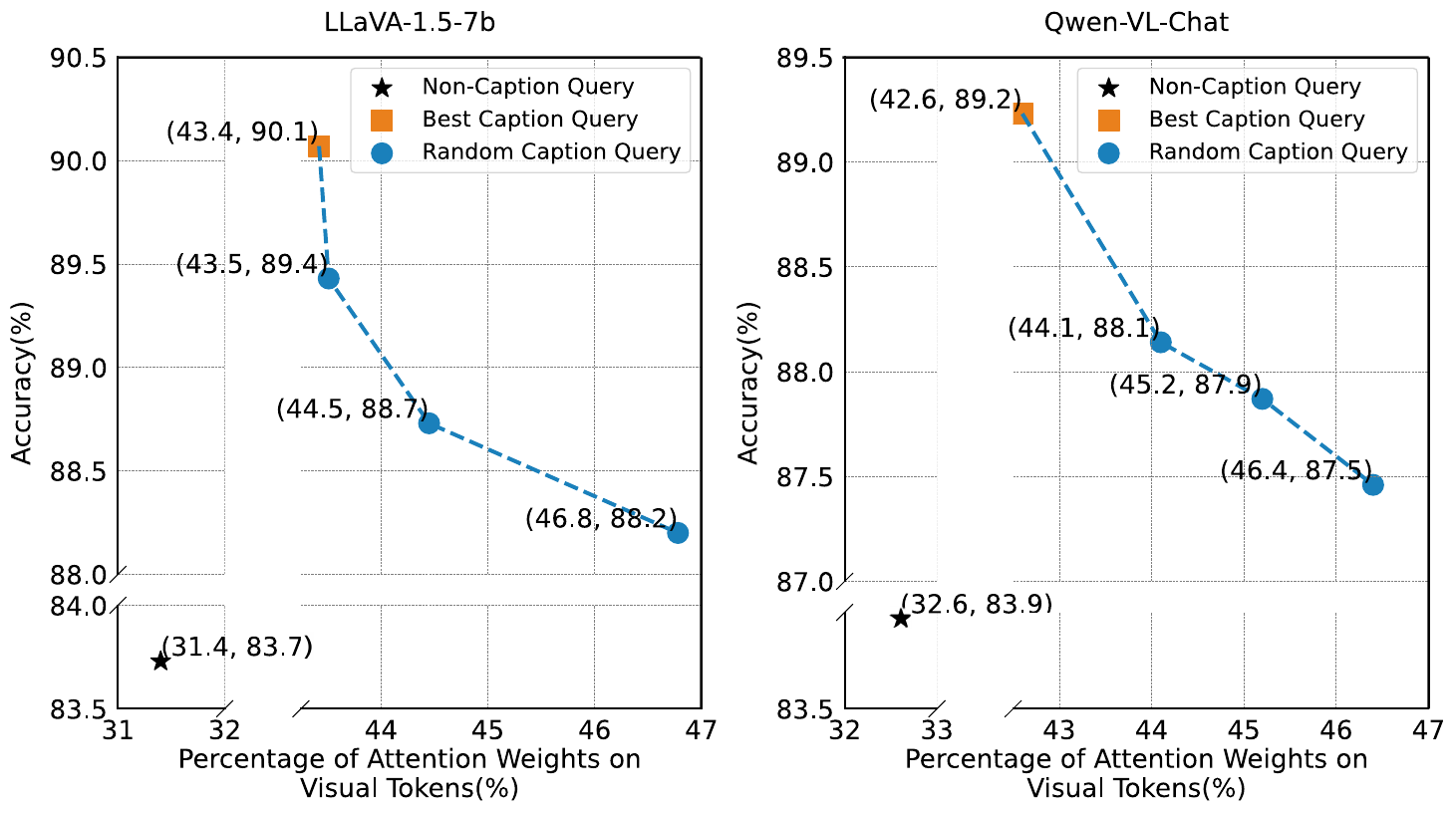}
  \caption{The accuracies of baselines and CAI with different caption queries on GQA Random POPE task.   }
  \label{fig:select}
\end{figure}
 
\noindent To better understand the necessity of the search algorithm, we will focus on analyzing the following two issues:

One potential question is,
\textit{Why does CAI prefer using a single caption query instead of combining multiple queries?} The answer lies in the fact that different caption queries activate different paths during inference. Combining multiple caption queries causes interference between the activated paths, preventing the performance improvements that could be achieved by using any single caption query.

Another key question is, \textit{Why does CAI select the query with minimal attention weight shift as the best?} The primary goal of CAI is to refine the outputs of caption-sensitive attention heads without significantly altering LVLMs' existing behaviors. By minimizing the attention weight shift, CAI strikes a balance between enhancing visual perception and maintaining the integrity of other foundational capabilities.
As illustrated in Figure 5, the experimental results demonstrate that LVLMs achieve the best performance using CAI search strategy. Certain caption query without careful selection may lead LVLMs to excessively focus on visual information, preventing it from achieving maximum performance improvement.

\subsection{Distribution of Probed Attention Heads}
\begin{figure}[!ht]
  \includegraphics[width=\columnwidth]{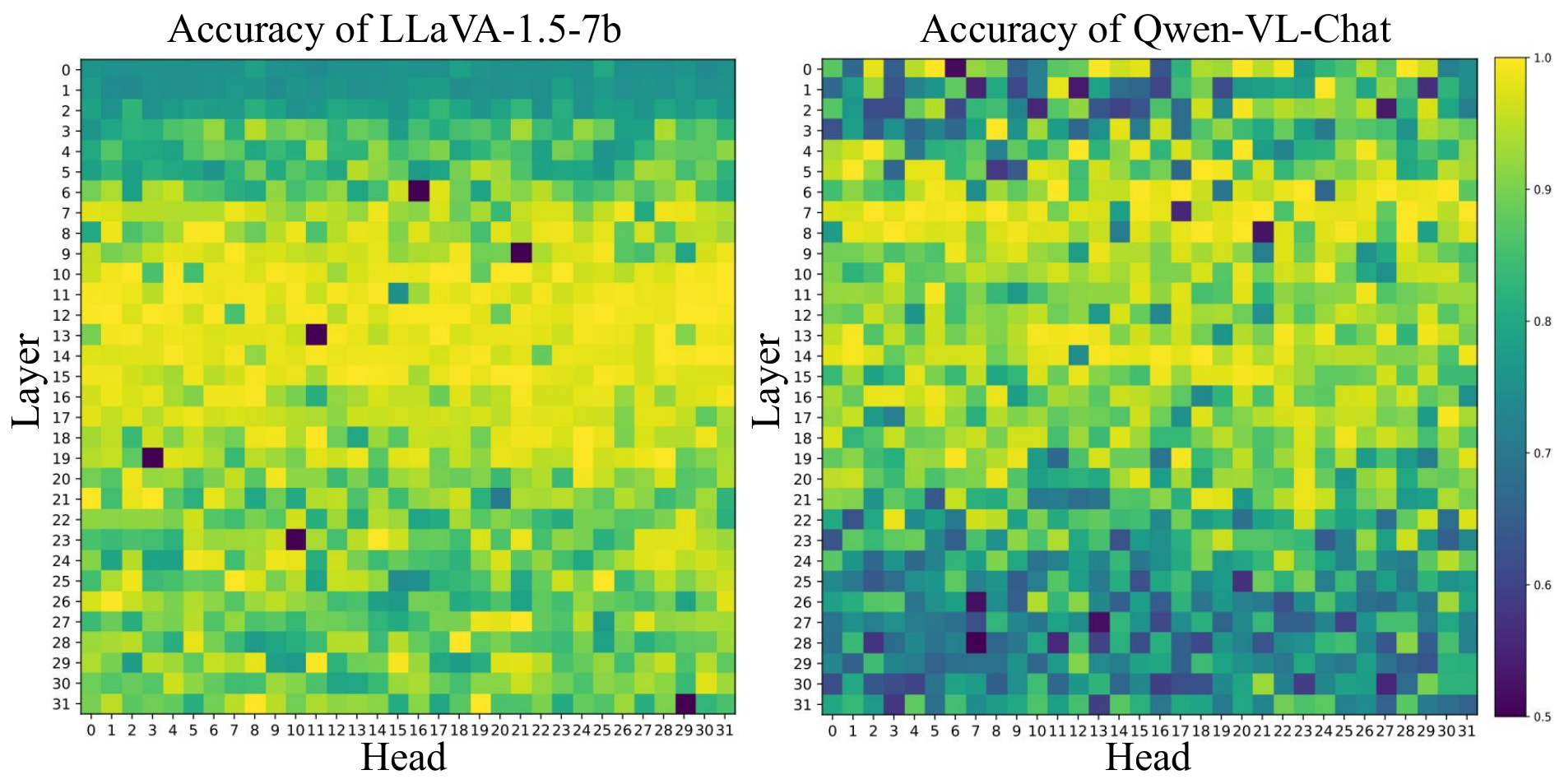}
  \caption{The accuracies of classifiers. 
  }
  \label{fig:heatmap}
\end{figure}

\begin{figure}[!ht]
  \includegraphics[width=\columnwidth]{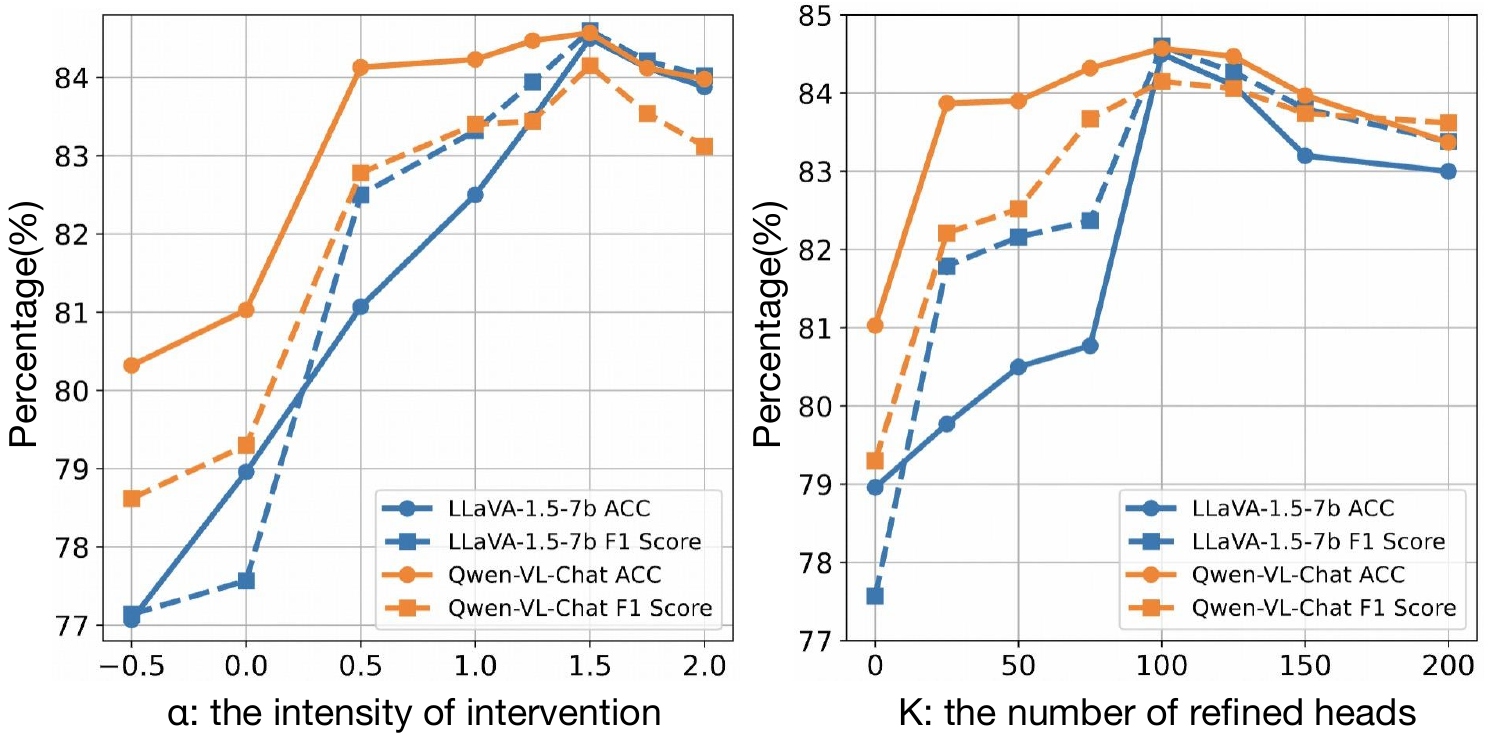}
  \caption{Ablation study of $\alpha$ and $K$. }
  \label{fig:abalation}
\end{figure}

As illustrated in Figure \ref{fig:heatmap}, we visualize the classification accuracies across 32 × 32 attention heads.
We observe that caption-sensitive attention heads are concentrated primarily between the 7th and 20th layers, which provides corroborating evidence for the quantization experiments detailed in Figure \ref{fig:quan}. These layers are critical for balancing visual perception and semantic understanding within the model. By refining the output of these attention heads, CAI significantly enhances LVLMs' visual perception capability and mitigate hallucination.

\subsection{Implications of Hyperparameters}

CAI method primarily relies on two key hyperparameters: the intensity of intervention $\alpha$ and the number of refined attention heads $K$. 
We performed a series of ablation experiments using greedy decoding on the MS-COCO Adversarial POPE dataset. As shown in Figure \ref{fig:abalation}, the key implications can be summarized as follows:

(1) Impact of $\alpha$: A negative value reduces the model's attention to visual tokens, which in turn diminishes its performance in hallucination mitigation. When $\alpha$ is small, the attention intervention is insufficient, resulting in only marginal improvements in model performance. A large $\alpha$ leads to insufficient attention to textual information, resulting in a decline in performance.

\begin{figure*}[!ht]
  \includegraphics[width=1.0\linewidth]{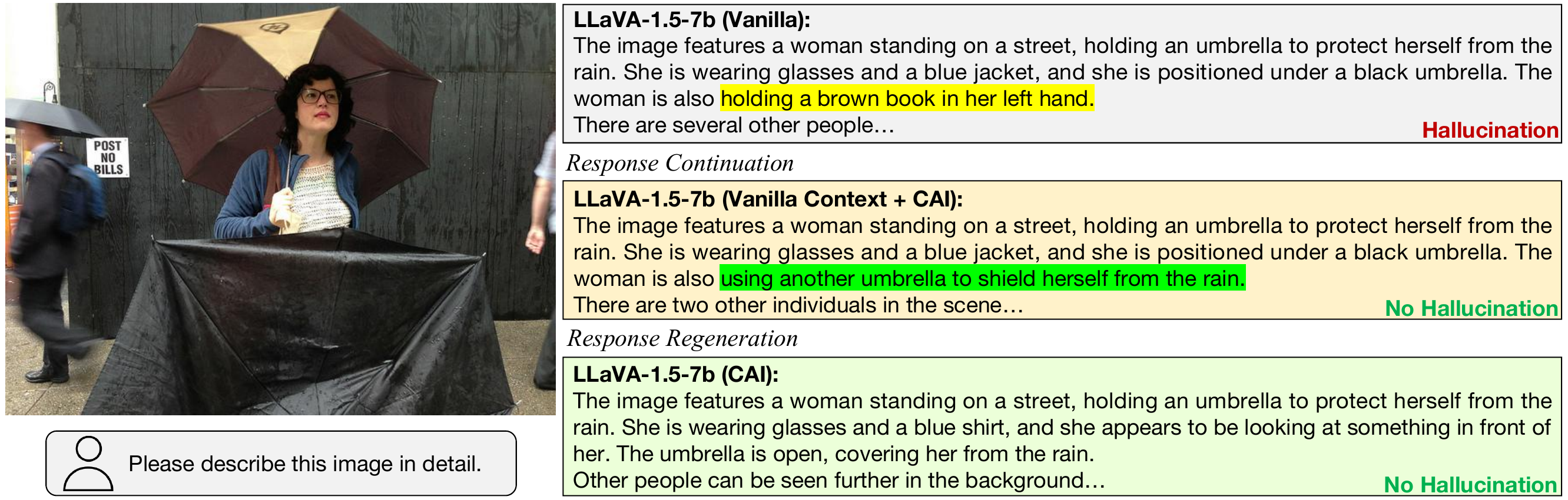}
  \caption{LLaVA-1.5-7b demonstrated hallucination when addressing caption query of MS-COCO CHAIR dataset. Both response continuation and regeneration for the same query provided by CAI effectively mitigate hallucination.}
  \label{fig:case_study}
\end{figure*}

(2) Impact of $K$: Applying intervention to too few attention heads fails to effectively influence the activation pathways of visual information, resulting in limited improvements in hallucination mitigation. Conversely, intervening in too many attention heads can disrupt critical attention activation paths that are unrelated to visual perception and play essential roles in other foundational capabilities, leading to performance degradation. 

See Appendix \ref{sec:E} for impact of over-intervention. 

\subsection{Case Study on CHAIR}
CAI proves effective in mitigating hallucination even when addressing caption queries. 
CAI strengthens the visual information attention of attention heads responsible for captioning tasks, further enhancing LVLMs' captioning capability.
As shown in Figure \ref{fig:case_study}, CAI effectively mitigates hallucination through sentence continuation and regeneration. 
See the Appendix \ref{sec:H} for more cases.

\section{Related Works}

\subsection{Large Vision-Language Models}
Several powerful LVLMs based on open-source LLM backbones combined with visual encoders have achieved impressive capabilities through extensive vision-language pretraining. Furthermore, recent searches have further improved model performance by employing high-resolution visual encoders \cite{hong2024cogvlm2} and exploring reinforcement learning methods, such as RLHF \cite{yu2024rlhf}. Closed-source models, such as GPT-4o \cite{hurst2024gpt} and Gemini 1.5 \cite{reid2024gemini} have demonstrated even more powerful performance.
However, recent LVLMs still suffer from hallucination problems. Addressing how to cost-effectively mitigate hallucination is still a critical problem that requires further exploration.

\subsection{Mitigating Hallucination in LVLMs}
Current methods for mitigating hallucination in LVLMs can be broadly categorized into two types: data-driven training methods and training-free methods. Training-based methods typically involve introducing novel training objectives \cite{chen2024alleviating} and utilizing carefully curated datasets \cite{gunjal2024detecting,liu2023mitigating,yu2024rlaif,you2023ferret}. For training-free methodes, the main strategies include designing decoding techniques \cite{leng2024mitigating,chen2024halc,chuang2023dola,huang2024opera,zhong2024investigatingmitigatingmultimodalhallucination} during the inference phase, leveraging language or visual prompts \cite{lee2023volcano,an2024agla}, incorporating external tools or knowledge sources \cite{zhao2024mitigating} and correct the generation \cite{yin2024woodpecker}. 
Furthermore, a key approach to reduce hallucination is to address attention deficits by adjusting or using decoding strategies \cite{an2024agla,gong2024damro,xing2024mitigating}. PAI \cite{liu2024paying} intervenes in attention heads by leveraging the direction and magnitude of their original outputs, and optimizes the output distribution during decoding to mitigate hallucinations. VTI \cite{liu2024reducing} reduces hallucinations by steering layer-level latent space representations during inference to enhance the stability of vision features.
However, our work is the first to explicitly reveal the impact of caption vs. non-caption queries on the attention activation patterns of LVLMs and mitigate hallucination by applying caption-sensitive attention intervention at head level during the inference.

\section{Conclusion}
In this paper, we presented CAI, a training-free method that refines caption-sensitive attention heads outputs for non-caption queries towards outputs for caption queries during the inference, thereby leveraging LVLMs' inherent fine-grained visual perception capabilities to mitigate object hallucination. 
CAI demonstrates strong generalizability and can be applied to several open-source LVLMs. 
Consistent performance improvements across diverse benchmarks highlight its robustness. 

\section{Limitations}
While CAI demonstrates significant effectiveness in mitigating object hallucinations in LVLMs, several limitations should be acknowledged to provide a balanced perspective on its applicability and scope. First, CAI relies on the availability of suitable caption queries to identify the optimal attention shift. Expanding the caption query candidate list could address this issue. Moreover, LVLMs that do not rely on multi-head attention mechanisms, or those employing non-standard visual-textual alignment strategies, may not benefit from CAI's intervention method. We will address the above issues in future work.


\bibliography{custom}

\begin{thebibliography}{47}
\providecommand{\natexlab}[1]{#1}

\bibitem[{An et~al.(2024)An, Tian, Leng, Nie, Lin, Wang, Dai, Chen, and Lu}]{an2024agla}
Wenbin An, Feng Tian, Sicong Leng, Jiahao Nie, Haonan Lin, QianYing Wang, Guang Dai, Ping Chen, and Shijian Lu. 2024.
\newblock Agla: Mitigating object hallucinations in large vision-language models with assembly of global and local attention.
\newblock \emph{arXiv preprint arXiv:2406.12718}.

\bibitem[{Arif et~al.(2025)Arif, Dip, Hussain, Zhang, and Thomas}]{arif2025fixing}
Kazi Hasan~Ibn Arif, Sajib~Acharjee Dip, Khizar Hussain, Lang Zhang, and Chris Thomas. 2025.
\newblock Fixing imbalanced attention to mitigate in-context hallucination of large vision-language model.
\newblock \emph{arXiv preprint arXiv:2501.12206}.

\bibitem[{Bai et~al.(2023)Bai, Bai, Yang, Wang, Tan, Wang, Lin, Zhou, and Zhou}]{bai2023qwen}
Jinze Bai, Shuai Bai, Shusheng Yang, Shijie Wang, Sinan Tan, Peng Wang, Junyang Lin, Chang Zhou, and Jingren Zhou. 2023.
\newblock Qwen-vl: A frontier large vision-language model with versatile abilities.
\newblock \emph{arXiv preprint arXiv:2308.12966}.

\bibitem[{Bi et~al.(2024{\natexlab{a}})Bi, Guo, Tang, Wen, Liu, and Xu}]{Bi2024UnveilingVP}
Jing Bi, Junjia Guo, Yunlong Tang, Lianggong Wen, Zhang Liu, and Chenliang Xu. 2024{\natexlab{a}}.
\newblock \href {https://api.semanticscholar.org/CorpusID:274992055} {Unveiling visual perception in language models: An attention head analysis approach}.
\newblock \emph{ArXiv}, abs/2412.18108.

\bibitem[{Bi et~al.(2024{\natexlab{b}})Bi, Guo, Tang, Wen, Liu, and Xu}]{bi2024unveiling}
Jing Bi, Junjia Guo, Yunlong Tang, Lianggong~Bruce Wen, Zhang Liu, and Chenliang Xu. 2024{\natexlab{b}}.
\newblock Unveiling visual perception in language models: An attention head analysis approach.
\newblock \emph{arXiv preprint arXiv:2412.18108}.

\bibitem[{Chen et~al.(2024{\natexlab{a}})Chen, Lyu, Gao, Song, and Shen}]{chen2024alleviating}
Beitao Chen, Xinyu Lyu, Lianli Gao, Jingkuan Song, and Heng~Tao Shen. 2024{\natexlab{a}}.
\newblock Alleviating hallucinations in large vision-language models through hallucination-induced optimization.
\newblock \emph{arXiv preprint arXiv:2405.15356}.

\bibitem[{Chen et~al.(2024{\natexlab{b}})Chen, Zhao, Luo, Yao, Li, and Zhou}]{chen2024halc}
Zhaorun Chen, Zhuokai Zhao, Hongyin Luo, Huaxiu Yao, Bo~Li, and Jiawei Zhou. 2024{\natexlab{b}}.
\newblock Halc: Object hallucination reduction via adaptive focal-contrast decoding.
\newblock \emph{arXiv preprint arXiv:2403.00425}.

\bibitem[{Chuang et~al.(2023)Chuang, Xie, Luo, Kim, Glass, and He}]{chuang2023dola}
Yung-Sung Chuang, Yujia Xie, Hongyin Luo, Yoon Kim, James Glass, and Pengcheng He. 2023.
\newblock Dola: Decoding by contrasting layers improves factuality in large language models.
\newblock \emph{arXiv preprint arXiv:2309.03883}.

\bibitem[{Cortes(1995)}]{cortes1995support}
Corinna Cortes. 1995.
\newblock Support-vector networks.
\newblock \emph{Machine Learning}.

\bibitem[{Favero et~al.(2024)Favero, Zancato, Trager, Choudhary, Perera, Achille, Swaminathan, and Soatto}]{favero2024multi}
Alessandro Favero, Luca Zancato, Matthew Trager, Siddharth Choudhary, Pramuditha Perera, Alessandro Achille, Ashwin Swaminathan, and Stefano Soatto. 2024.
\newblock Multi-modal hallucination control by visual information grounding.
\newblock In \emph{Proceedings of the IEEE/CVF Conference on Computer Vision and Pattern Recognition}, pages 14303--14312.

\bibitem[{Fu et~al.(2023)Fu, Chen, Shen, Qin, Zhang, Lin, Qiu, Lin, Yang, Zheng, Li, Sun, and Ji}]{Fu2023MMEAC}
Chaoyou Fu, Peixian Chen, Yunhang Shen, Yulei Qin, Mengdan Zhang, Xu~Lin, Zhenyu Qiu, Wei Lin, Jinrui Yang, Xiawu Zheng, Ke~Li, Xing Sun, and Rongrong Ji. 2023.
\newblock \href {https://api.semanticscholar.org/CorpusID:259243928} {Mme: A comprehensive evaluation benchmark for multimodal large language models}.
\newblock \emph{ArXiv}, abs/2306.13394.

\bibitem[{Ghosh et~al.(2024)Ghosh, Evuru, Kumar, Tyagi, Nieto, Jin, and Manocha}]{ghosh2024vdgdmitigatinglvlmhallucinations}
Sreyan Ghosh, Chandra Kiran~Reddy Evuru, Sonal Kumar, Utkarsh Tyagi, Oriol Nieto, Zeyu Jin, and Dinesh Manocha. 2024.
\newblock \href {https://arxiv.org/abs/2405.15683} {Vdgd: Mitigating lvlm hallucinations in cognitive prompts by bridging the visual perception gap}.
\newblock \emph{Preprint}, arXiv:2405.15683.

\bibitem[{Gong et~al.(2024)Gong, Ming, Wang, and Wei}]{gong2024damro}
Xuan Gong, Tianshi Ming, Xinpeng Wang, and Zhihua Wei. 2024.
\newblock Damro: Dive into the attention mechanism of lvlm to reduce object hallucination.
\newblock \emph{arXiv preprint arXiv:2410.04514}.

\bibitem[{Gunjal et~al.(2024)Gunjal, Yin, and Bas}]{gunjal2024detecting}
Anisha Gunjal, Jihan Yin, and Erhan Bas. 2024.
\newblock Detecting and preventing hallucinations in large vision language models.
\newblock In \emph{Proceedings of the AAAI Conference on Artificial Intelligence}, volume~38, pages 18135--18143.

\bibitem[{Hong et~al.(2024)Hong, Wang, Ding, Yu, Lv, Wang, Cheng, Huang, Ji, Xue et~al.}]{hong2024cogvlm2}
Wenyi Hong, Weihan Wang, Ming Ding, Wenmeng Yu, Qingsong Lv, Yan Wang, Yean Cheng, Shiyu Huang, Junhui Ji, Zhao Xue, et~al. 2024.
\newblock Cogvlm2: Visual language models for image and video understanding.
\newblock \emph{arXiv preprint arXiv:2408.16500}.

\bibitem[{Huang et~al.(2023)Huang, Yu, Ma, Zhong, Feng, Wang, Chen, Peng, Feng, Qin et~al.}]{huang2023survey}
Lei Huang, Weijiang Yu, Weitao Ma, Weihong Zhong, Zhangyin Feng, Haotian Wang, Qianglong Chen, Weihua Peng, Xiaocheng Feng, Bing Qin, et~al. 2023.
\newblock A survey on hallucination in large language models: Principles, taxonomy, challenges, and open questions.
\newblock \emph{arXiv preprint arXiv:2311.05232}.

\bibitem[{Huang et~al.(2024)Huang, Dong, Zhang, Wang, He, Wang, Lin, Zhang, and Yu}]{huang2024opera}
Qidong Huang, Xiaoyi Dong, Pan Zhang, Bin Wang, Conghui He, Jiaqi Wang, Dahua Lin, Weiming Zhang, and Nenghai Yu. 2024.
\newblock Opera: Alleviating hallucination in multi-modal large language models via over-trust penalty and retrospection-allocation.
\newblock In \emph{Proceedings of the IEEE/CVF Conference on Computer Vision and Pattern Recognition}, pages 13418--13427.

\bibitem[{Hurst et~al.(2024)Hurst, Lerer, Goucher, Perelman, Ramesh, Clark, Ostrow, Welihinda, Hayes, Radford et~al.}]{hurst2024gpt}
Aaron Hurst, Adam Lerer, Adam~P Goucher, Adam Perelman, Aditya Ramesh, Aidan Clark, AJ~Ostrow, Akila Welihinda, Alan Hayes, Alec Radford, et~al. 2024.
\newblock Gpt-4o system card.
\newblock \emph{arXiv preprint arXiv:2410.21276}.

\bibitem[{Kan et~al.(2024)Kan, Zhang, Liao, Tian, Yang, Xiao, Li, Jiang, Wang, and Liao}]{kan2024catch}
Zhehan Kan, Ce~Zhang, Zihan Liao, Yapeng Tian, Wenming Yang, Junyuan Xiao, Xu~Li, Dongmei Jiang, Yaowei Wang, and Qingmin Liao. 2024.
\newblock Catch: Complementary adaptive token-level contrastive decoding to mitigate hallucinations in lvlms.
\newblock \emph{arXiv preprint arXiv:2411.12713}.

\bibitem[{Lau et~al.(2018)Lau, Gayen, Ben~Abacha, and Demner-Fushman}]{lau2018dataset}
Jason~J Lau, Soumya Gayen, Asma Ben~Abacha, and Dina Demner-Fushman. 2018.
\newblock A dataset of clinically generated visual questions and answers about radiology images.
\newblock \emph{Scientific data}, 5(1):1--10.

\bibitem[{Lee et~al.(2023)Lee, Park, Jo, and Seo}]{lee2023volcano}
Seongyun Lee, Sue~Hyun Park, Yongrae Jo, and Minjoon Seo. 2023.
\newblock Volcano: mitigating multimodal hallucination through self-feedback guided revision.
\newblock \emph{arXiv preprint arXiv:2311.07362}.

\bibitem[{Leng et~al.(2024)Leng, Zhang, Chen, Li, Lu, Miao, and Bing}]{leng2024mitigating}
Sicong Leng, Hang Zhang, Guanzheng Chen, Xin Li, Shijian Lu, Chunyan Miao, and Lidong Bing. 2024.
\newblock Mitigating object hallucinations in large vision-language models through visual contrastive decoding.
\newblock In \emph{Proceedings of the IEEE/CVF Conference on Computer Vision and Pattern Recognition}, pages 13872--13882.

\bibitem[{Li et~al.(2025)Li, Zhang, Jie, Ma, and Li}]{li2025mitigating}
Jiaming Li, Jiacheng Zhang, Zequn Jie, Lin Ma, and Guanbin Li. 2025.
\newblock Mitigating hallucination for large vision language model by inter-modality correlation calibration decoding.
\newblock \emph{arXiv preprint arXiv:2501.01926}.

\bibitem[{Li et~al.(2024)Li, Patel, Vi{\'e}gas, Pfister, and Wattenberg}]{li2024inference}
Kenneth Li, Oam Patel, Fernanda Vi{\'e}gas, Hanspeter Pfister, and Martin Wattenberg. 2024.
\newblock Inference-time intervention: Eliciting truthful answers from a language model.
\newblock \emph{Advances in Neural Information Processing Systems}, 36.

\bibitem[{Li et~al.(2023)Li, Du, Zhou, Wang, Zhao, and Wen}]{li2023evaluating}
Yifan Li, Yifan Du, Kun Zhou, Jinpeng Wang, Wayne~Xin Zhao, and Ji-Rong Wen. 2023.
\newblock Evaluating object hallucination in large vision-language models.
\newblock \emph{arXiv preprint arXiv:2305.10355}.

\bibitem[{Lin et~al.(2014)Lin, Maire, Belongie, Hays, Perona, Ramanan, Doll{\'a}r, and Zitnick}]{lin2014microsoft}
Tsung-Yi Lin, Michael Maire, Serge Belongie, James Hays, Pietro Perona, Deva Ramanan, Piotr Doll{\'a}r, and C~Lawrence Zitnick. 2014.
\newblock Microsoft coco: Common objects in context.
\newblock In \emph{Computer Vision--ECCV 2014: 13th European Conference, Zurich, Switzerland, September 6-12, 2014, Proceedings, Part V 13}, pages 740--755. Springer.

\bibitem[{Liu et~al.(2023)Liu, Lin, Li, Wang, Yacoob, and Wang}]{liu2023mitigating}
Fuxiao Liu, Kevin Lin, Linjie Li, Jianfeng Wang, Yaser Yacoob, and Lijuan Wang. 2023.
\newblock Mitigating hallucination in large multi-modal models via robust instruction tuning.
\newblock In \emph{The Twelfth International Conference on Learning Representations}.

\bibitem[{Liu et~al.(2024{\natexlab{a}})Liu, Li, Li, and Lee}]{liu2024improved}
Haotian Liu, Chunyuan Li, Yuheng Li, and Yong~Jae Lee. 2024{\natexlab{a}}.
\newblock Improved baselines with visual instruction tuning.
\newblock In \emph{Proceedings of the IEEE/CVF Conference on Computer Vision and Pattern Recognition}, pages 26296--26306.

\bibitem[{Liu et~al.(2024{\natexlab{b}})Liu, Li, Li, Li, Zhang, Shen, and Lee}]{liu2024llavanext}
Haotian Liu, Chunyuan Li, Yuheng Li, Bo~Li, Yuanhan Zhang, Sheng Shen, and Yong~Jae Lee. 2024{\natexlab{b}}.
\newblock \href {https://llava-vl.github.io/blog/2024-01-30-llava-next/} {Llava-next: Improved reasoning, ocr, and world knowledge}.

\bibitem[{Liu et~al.(2024{\natexlab{c}})Liu, Ye, and Zou}]{liu2024reducing}
Sheng Liu, Haotian Ye, and James Zou. 2024{\natexlab{c}}.
\newblock Reducing hallucinations in vision-language models via latent space steering.
\newblock \emph{arXiv preprint arXiv:2410.15778}.

\bibitem[{Liu et~al.(2024{\natexlab{d}})Liu, Zheng, and Chen}]{liu2024paying}
Shi Liu, Kecheng Zheng, and Wei Chen. 2024{\natexlab{d}}.
\newblock Paying more attention to image: A training-free method for alleviating hallucination in lvlms.
\newblock \emph{arXiv preprint arXiv:2407.21771}.

\bibitem[{Liu et~al.(2024{\natexlab{e}})Liu, Duan, Zhang, Li, Zhang, Zhao, Yuan, Wang, He, Liu et~al.}]{liu2024mmbench}
Yuan Liu, Haodong Duan, Yuanhan Zhang, Bo~Li, Songyang Zhang, Wangbo Zhao, Yike Yuan, Jiaqi Wang, Conghui He, Ziwei Liu, et~al. 2024{\natexlab{e}}.
\newblock Mmbench: Is your multi-modal model an all-around player?
\newblock In \emph{European conference on computer vision}, pages 216--233. Springer.

\bibitem[{Reid et~al.(2024)Reid, Savinov, Teplyashin, Lepikhin, Lillicrap, Alayrac, Soricut, Lazaridou, Firat, Schrittwieser et~al.}]{reid2024gemini}
Machel Reid, Nikolay Savinov, Denis Teplyashin, Dmitry Lepikhin, Timothy Lillicrap, Jean-baptiste Alayrac, Radu Soricut, Angeliki Lazaridou, Orhan Firat, Julian Schrittwieser, et~al. 2024.
\newblock Gemini 1.5: Unlocking multimodal understanding across millions of tokens of context.
\newblock \emph{arXiv preprint arXiv:2403.05530}.

\bibitem[{Rohrbach et~al.(2018)Rohrbach, Hendricks, Burns, Darrell, and Saenko}]{rohrbach2018object}
Anna Rohrbach, Lisa~Anne Hendricks, Kaylee Burns, Trevor Darrell, and Kate Saenko. 2018.
\newblock Object hallucination in image captioning.
\newblock \emph{arXiv preprint arXiv:1809.02156}.

\bibitem[{Sahoo et~al.(2024)Sahoo, Meharia, Ghosh, Saha, Jain, and Chadha}]{sahoo2024comprehensive}
Pranab Sahoo, Prabhash Meharia, Akash Ghosh, Sriparna Saha, Vinija Jain, and Aman Chadha. 2024.
\newblock A comprehensive survey of hallucination in large language, image, video and audio foundation models.
\newblock \emph{Findings of the Association for Computational Linguistics: EMNLP 2024}, pages 11709--11724.

\bibitem[{Sun et~al.(2023)Sun, Shen, Cao, Liu, Li, Shen, Gan, Gui, Wang, Yang et~al.}]{sun2023aligning}
Zhiqing Sun, Sheng Shen, Shengcao Cao, Haotian Liu, Chunyuan Li, Yikang Shen, Chuang Gan, Liang-Yan Gui, Yu-Xiong Wang, Yiming Yang, et~al. 2023.
\newblock Aligning large multimodal models with factually augmented rlhf.
\newblock \emph{arXiv preprint arXiv:2309.14525}.

\bibitem[{Wang et~al.(2024)Wang, Pan, Ding, and Biemann}]{wang2024mitigating}
Xintong Wang, Jingheng Pan, Liang Ding, and Chris Biemann. 2024.
\newblock Mitigating hallucinations in large vision-language models with instruction contrastive decoding.
\newblock \emph{arXiv preprint arXiv:2403.18715}.

\bibitem[{Xing et~al.(2024)Xing, Li, Laptev, and Lu}]{xing2024mitigating}
Yun Xing, Yiheng Li, Ivan Laptev, and Shijian Lu. 2024.
\newblock Mitigating object hallucination via concentric causal attention.
\newblock \emph{arXiv preprint arXiv:2410.15926}.

\bibitem[{Yin et~al.(2024)Yin, Fu, Zhao, Xu, Wang, Sui, Shen, Li, Sun, and Chen}]{yin2024woodpecker}
Shukang Yin, Chaoyou Fu, Sirui Zhao, Tong Xu, Hao Wang, Dianbo Sui, Yunhang Shen, Ke~Li, Xing Sun, and Enhong Chen. 2024.
\newblock Woodpecker: Hallucination correction for multimodal large language models.
\newblock \emph{Science China Information Sciences}, 67(12):220105.

\bibitem[{You et~al.(2023)You, Zhang, Gan, Du, Zhang, Wang, Cao, Chang, and Yang}]{you2023ferret}
Haoxuan You, Haotian Zhang, Zhe Gan, Xianzhi Du, Bowen Zhang, Zirui Wang, Liangliang Cao, Shih-Fu Chang, and Yinfei Yang. 2023.
\newblock Ferret: Refer and ground anything anywhere at any granularity.
\newblock \emph{arXiv preprint arXiv:2310.07704}.

\bibitem[{Yu et~al.(2024{\natexlab{a}})Yu, Yao, Zhang, He, Han, Cui, Hu, Liu, Zheng, Sun et~al.}]{yu2024rlhf}
Tianyu Yu, Yuan Yao, Haoye Zhang, Taiwen He, Yifeng Han, Ganqu Cui, Jinyi Hu, Zhiyuan Liu, Hai-Tao Zheng, Maosong Sun, et~al. 2024{\natexlab{a}}.
\newblock Rlhf-v: Towards trustworthy mllms via behavior alignment from fine-grained correctional human feedback.
\newblock In \emph{Proceedings of the IEEE/CVF Conference on Computer Vision and Pattern Recognition}, pages 13807--13816.

\bibitem[{Yu et~al.(2024{\natexlab{b}})Yu, Zhang, Yao, Dang, Chen, Lu, Cui, He, Liu, Chua et~al.}]{yu2024rlaif}
Tianyu Yu, Haoye Zhang, Yuan Yao, Yunkai Dang, Da~Chen, Xiaoman Lu, Ganqu Cui, Taiwen He, Zhiyuan Liu, Tat-Seng Chua, et~al. 2024{\natexlab{b}}.
\newblock Rlaif-v: Aligning mllms through open-source ai feedback for super gpt-4v trustworthiness.
\newblock \emph{arXiv preprint arXiv:2405.17220}.

\bibitem[{Zhao et~al.(2024)Zhao, Deng, Zhang, and Gu}]{zhao2024mitigating}
Linxi Zhao, Yihe Deng, Weitong Zhang, and Quanquan Gu. 2024.
\newblock Mitigating object hallucination in large vision-language models via classifier-free guidance.
\newblock \emph{arXiv preprint arXiv:2402.08680}.

\bibitem[{Zhong et~al.(2022)Zhong, Liu, Yin, Mao, Jiao, Liu, Zhu, Ji, and Han}]{zhong2022towards}
Ming Zhong, Yang Liu, Da~Yin, Yuning Mao, Yizhu Jiao, Pengfei Liu, Chenguang Zhu, Heng Ji, and Jiawei Han. 2022.
\newblock Towards a unified multi-dimensional evaluator for text generation.
\newblock \emph{arXiv preprint arXiv:2210.07197}.

\bibitem[{Zhong et~al.(2024)Zhong, Feng, Zhao, Li, Huang, Gu, Ma, Xu, and Qin}]{zhong2024investigatingmitigatingmultimodalhallucination}
Weihong Zhong, Xiaocheng Feng, Liang Zhao, Qiming Li, Lei Huang, Yuxuan Gu, Weitao Ma, Yuan Xu, and Bing Qin. 2024.
\newblock \href {https://arxiv.org/abs/2407.00569} {Investigating and mitigating the multimodal hallucination snowballing in large vision-language models}.
\newblock \emph{Preprint}, arXiv:2407.00569.

\bibitem[{Zhou et~al.(2024)Zhou, Yan, Zou, Wang, Liu, and Hu}]{zhou2024mitigating}
Guanyu Zhou, Yibo Yan, Xin Zou, Kun Wang, Aiwei Liu, and Xuming Hu. 2024.
\newblock Mitigating modality prior-induced hallucinations in multimodal large language models via deciphering attention causality.
\newblock \emph{arXiv preprint arXiv:2410.04780}.

\bibitem[{Zhu et~al.(2024)Zhu, Ji, Chen, Xu, Ye, and Liu}]{zhu2024ibd}
Lanyun Zhu, Deyi Ji, Tianrun Chen, Peng Xu, Jieping Ye, and Jun Liu. 2024.
\newblock Ibd: Alleviating hallucinations in large vision-language models via image-biased decoding.
\newblock \emph{arXiv preprint arXiv:2402.18476}.

\end{thebibliography}

\clearpage
\appendix

\section{Additional Experimental Details}
\label{sec:A}
All datasets used in this paper are licensed under a \href{https://creativecommons.org/licenses/by/4.0/legalcode}{Creative Commons Attribution 4.0 License}.
\subsection{Data Source}
Although our method does not rely on specific data, we separately specify the sources of the data used in the experiments for the sake of reproducibility.
\subsubsection{Data of Best Query Search}
In the best caption search algorithm, we use the top 100 VQA samples from the complex reasoning data in the LLaVA-1.5-7b pre-training dataset. From this, we obtain non-caption queries and their corresponding images. Additionally, we maintain a list of 13 candidate caption queries, some of which are manually generated and others are derived from the pre-trained instructions of LLaVA-1.5-7b. The caption query candidates are listed as follows:

\textit{"What do you see happening in this image?"}, 
\textit{"What do you think is going on in this snapshot?"},
\textit{"Can you elaborate on the elements of the picture provided?"},
\textit{"Describe the following image."},
\textit{"What's happening in the scene?"},
\textit{"Analyze the image in a comprehensive and detailed manner."},
\textit{"Write a detailed description of the given image."},
\textit{"What is this photo about?"},
\textit{"Explain the visual content of the image in great detail."},
\textit{"What are the key elements in this picture?"},
\textit{"Can you describe the main features of this image for me?"},
\textit{"Please describe this image in detail."},
\textit{"Generate the caption in English:"}

In the experiments, the best caption query for LLaVA-1.5-7b is \textit{"Analyze the image in a comprehensive and detailed manner."} and the best caption query for Qwen-VL-Chat is \textit{"Please describe this image in detail."}.
\subsubsection{Data of Probe and Shift Computation }
We extracted the first 1,000 samples from the complex reasoning data in the LLaVA-1.5-7b pre-training dataset. The questions from these samples were treated as non-caption queries.
\subsection{Detailed Experimental Setup}
In the experiment of POPE, 'regular' refers to the direct sampling setting. We used direct sampling decoding and set $\alpha=1.5$ and $K = 100$ in the main experiments.

\section{Comparison with More Advanced Methods}
\label{sec:B}
We selected LLaVA-1.5-7b as the baseline model and compared CAI with more advanced models including VCD \cite{leng2024mitigating}, ICD \cite{wang2024mitigating}, OPERA \cite{huang2024opera}, Woodpecker \cite{yin2024woodpecker}, M3ID \cite{favero2024multi}, DAMRO \cite{gong2024damro}, IMCCD \cite{li2025mitigating}, CATCH \cite{kan2024catch}, IBD \cite{zhu2024ibd} and CAUSALMM \cite{zhou2024mitigating}.
The results of CAI compared with several state-of-the-art methods on MS-COCO Adversarial POPE are shown in Table \ref{add_coco_result}.

\section{Detailed Experimental Results of MME}
\label{sec:C}
Detailed experimental results on MME perception and cognition can be found in Table \ref{add_mme_p_result} and Table \ref{add_mme_r_result}. 

\section{Detailed Experimental Results of Ablation}
\label{sec:D}
Detailed results of the ablation experiments can be found in Table \ref{add_ablation_a} and Table \ref{add_ablation_k}. 

\begin{table}[!ht]
\small
\centering
\renewcommand{\arraystretch}{1.2} 
\setlength{\tabcolsep}{3pt} 
\begin{tabular}{ccccc}
\toprule
 \multirow{2}{*}{ \textbf{$\alpha$}} & \multicolumn{2}{c}{\textbf{LLaVA-1.5-7b}} & \multicolumn{2}{c}{\textbf{Qwen-VL-Chat}}  \\
\cmidrule(lr){2-3}
\cmidrule(lr){4-5}
 &  Accuracy & F1-Score & Accuracy & F1-Score \\
\midrule
-0.50&	77.07&	77.14&	80.32&	78.62\\
0.00&	78.96&	77.57&	81.03&	79.30\\
0.50&	81.07&	82.50&	84.13&	82.78\\
1.00&	82.50&	83.32&	84.23&	83.40\\
1.25&	83.47&	83.94&	84.47&	83.44\\
1.50&	84.50&	84.60&	84.57&	84.15\\
1.75&	84.90&	84.21&	84.12&	83.54\\
2.00&	85.10&	84.02&	83.98&	83.12\\
\bottomrule
\end{tabular}
\caption{Detailed results of $\alpha$ on MS-COCO Adversarial POPE dataset. 
}

\label{add_ablation_a}
\end{table}

\begin{table}[!ht]
\small
\centering
\renewcommand{\arraystretch}{1.2} 
\setlength{\tabcolsep}{3pt} 
\begin{tabular}{ccccc}
\toprule
 \multirow{2}{*}{ \textbf{$K$}} & \multicolumn{2}{c}{\textbf{LLaVA-1.5-7b}} & \multicolumn{2}{c}{\textbf{Qwen-VL-Chat}}  \\
\cmidrule(lr){2-3}
\cmidrule(lr){4-5}
 &  Accuracy & F1-Score & Accuracy & F1-Score \\
\midrule
0	&78.96&	77.57&	81.03&	79.30\\
25	&79.77&	81.79&	83.87&	82.21\\
50	&80.50&	82.16&	83.90&	82.52\\
75&	80.77&	82.37&	84.32&	83.67\\
100&84.50&	84.60&	84.57&	84.15\\
125	&84.10&	84.27&	84.47&	84.06\\
150&83.20&	83.80&	83.97&	83.74\\
200	&83.00&	83.38&	83.37&	83.62\\
\bottomrule
\end{tabular}
\caption{Detailed results of $K$ on MS-COCO Adversarial POPE dataset. 
}
\label{add_ablation_k}
\end{table}

\begin{table*}[!ht]
\small
\centering
\renewcommand{\arraystretch}{1.2} 
\setlength{\tabcolsep}{3pt} 
\begin{tabular}{lcccccccc}
\toprule
\multirow{2}{*}{ \textbf{Method}} & \multicolumn{2}{c}{\textbf{Random}} & \multicolumn{2}{c}{\textbf{Popular}} & \multicolumn{2}{c}{\textbf{Adversarial}} & \multicolumn{2}{c}{\textbf{Average}}  \\
\cmidrule(lr){2-3}
\cmidrule(lr){4-5}
\cmidrule(lr){6-7}
\cmidrule(lr){8-9}
& Accuracy & F1-Score & Accuracy & F1-Score & Accuracy & F1-Score & Accuracy & F1-Score \\
\midrule
Regular &83.29  &81.33  &81.88  &80.06 &78.96 &77.57&81.38&79.65 \\
VCD \textcolor{gray}{\textit{(CVPR'24)}} &87.73 &87.16  &85.38  &85.06 & 80.88 & 81.33&84.66&84.52\\
ICD \textcolor{gray}{\textit{(EMNLP'24 findings)}}&89.56 &89.68 &86.16&86.76 &79.71 &81.70 &85.14 &86.05\\
OPERA \textcolor{gray}{\textit{(CVPR'24)}} &89.20  &88.81  &86.64  &86.62 &81.24 &81.38& 85.70& 85.60\\
Woodpecker \textcolor{gray}{\textit{(SCIS'24)}}&87.67 &86.45 &80.67 &79.72 &80.67 &80.00&83.00&82.05 \\
M3ID \textcolor{gray}{\textit{(CVPR'24)}} &86.20 &84.51 & 84.77& 83.17& 82.53 &81.14 &84.50&82.94 \\
DAMRO \textcolor{gray}{\textit{(EMNLP'24)}}&88.20 &87.29 & 85.67& 84.98 &82.07&81.90&85.31&84.72\\
IMCCD \textcolor{gray}{\textit{(arXiv'25)}}& 89.23 &88.68 &86.73 &86.13 &82.87 & 82.77&86.27&85.86\\
CATCH \textcolor{gray}{\textit{(ECCV'24)}} & \textbf{90.43} & \textbf{90.13} &87.07 &86.56 & 83.17& 83.18& 86.89 &86.62\\
VDD \textcolor{gray}{\textit{(arXiv'24)}}& 90.00 & 88.79 & 85.91& 84.40 & 83.52 &82.20&86.48&85.13\\
CAUSALMM \textcolor{gray}{\textit{(ICLR'25)}} & 88.93 &88.10 & 87.13 & 87.26 &83.70 &82.78&86.59&86.05 \\ 
\midrule
CAI(ours) & 89.87 & 89.43 &  \textbf{88.32} & \textbf{87.95} &\textbf{84.27} & \textbf{84.41} &\textbf{87.49} &\textbf{87.22}\\
\bottomrule
\end{tabular}
\caption{Result compared with more advanced methods on MS-COCO POPE. 
}
\label{add_coco_result}
\end{table*}

\begin{figure*}[!ht]
  \includegraphics[width=1.0\linewidth]{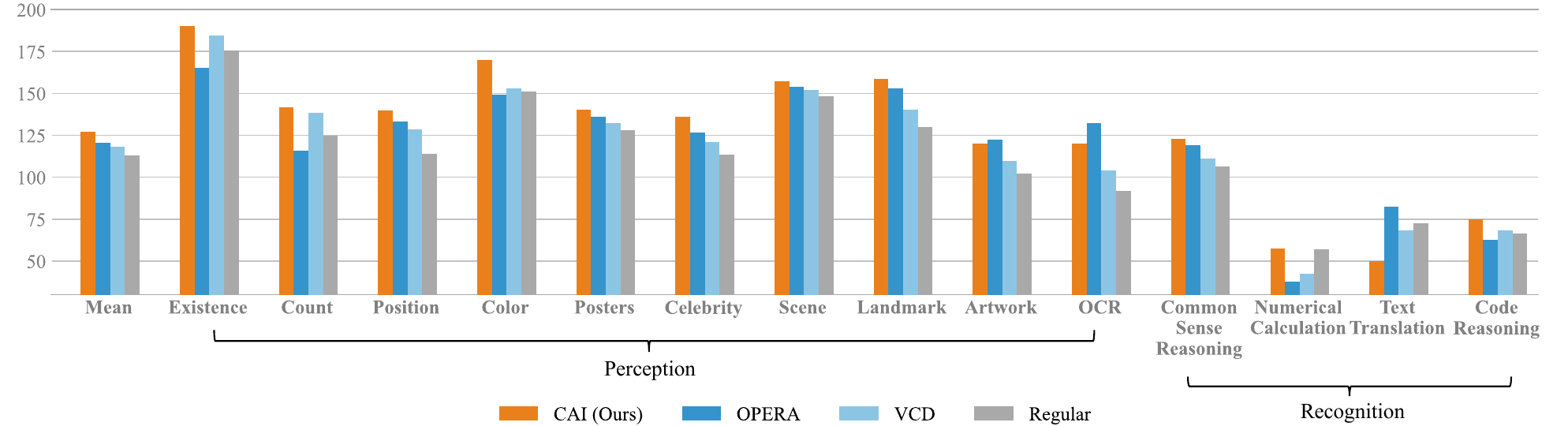}
  \caption{Main results on the MME full set. CAI leads to the best enhancement in LVLMs' perception capacities while preserving their recognition competencies.}
  \label{fig:mme_full}
\end{figure*}

\begin{table*}[!ht]
\small
\centering
\renewcommand{\arraystretch}{1.1} 
\setlength{\tabcolsep}{2pt} 
\begin{tabular}{lccccccccccc}
\toprule
Method & Artwork &Celebrity &Color &Count &Existence & Landmark &OCR &Position &Posters &Scene & Total \\
\midrule
Regular &102.20 &113.59 &151.00 &124.67 &175.67&129.95 &92.00 &114.00 &127.82 &148.30 &1279.20\\
VCD & 109.60&120.94&153.00&138.33&184.66&140.45&104.00&128.67&132.11&152.20&1363.96\\
OPERA &\textbf{122.50}&126.76&149.00&116.00&165.00&152.75&\textbf{132.50}&133.33&136.05&154.00&1387.89 \\
CAI(ours)&120.25 &\textbf{135.88} &\textbf{170.00} &\textbf{141.67}& \textbf{190.00}&\textbf{158.50}&120.00&\textbf{140.00} &\textbf{140.48}&\textbf{157.00}&\textbf{1473.78} \\
\bottomrule
\end{tabular}
\caption{Results on all MME perception-related tasks. The best performance of each setting is \textbf{bolded}. 
}
\label{add_mme_p_result}
\end{table*}

\begin{table*}[!ht]
\small
\centering
\renewcommand{\arraystretch}{1.1} 
\setlength{\tabcolsep}{3pt} 
\begin{tabular}{lccccc}
\toprule
Method & Coding Reasoning &Commonsense Reasoning & Numerical Calculation& Text Translation& Total \\
\midrule
Regular &66.38 &106.43&57.00&72.50&302.31 \\
VCD &68.50&111.29&42.64&68.50 &290.93 \\
OPERA &62.50 &119.29 &37.50 &\textbf{82.50}&301.79 \\
CAI(ours)& \textbf{75.00} &\textbf{122.86} &\textbf{57.50} &50.00&\textbf{305.36} \\
\bottomrule
\end{tabular}
\caption{Results on all MME recognition-related tasks. The best performance of each setting is \textbf{bolded}. 
}
\label{add_mme_r_result}
\end{table*}

\begin{table*}[!ht]
\small
\centering
\renewcommand{\arraystretch}{1.0} 
\setlength{\tabcolsep}{5pt} 
\begin{tabular}{lccccc}
\toprule
Method & CHAIRs & CHAIRi & PPL & Coherence & Fluency \\
\midrule
 Greedy & 20.80 & 6.77 & 3.97 & 0.998500 & 0.805269 \\
 CAI & 17.20 & 5.50 & 4.11 & 0.998352 & 0.791763 \\
 CAI (over-intervention) & 18.60 & 6.00 & 4.23 & 0.998180 & 0.809675 \\
\bottomrule
\end{tabular}
\caption{Impact of over-intervention on CHAIR}. 
\label{add_fluency_result}
\end{table*}

\section{Impact of Over-intervention}
\label{sec:E}
In this work, we have provided a detailed discussion of the probe for intervention heads, the number of intervention heads, and the intervention strength. Experimental results were used to determine various hyper-parameters to avoid over-intervention. Following prior work, we employ UniEval \cite{zhong2022towards} and perplexity (PPL) computation to assess the coherence and fluency. As the experimental results show in Table \ref{add_fluency_result}, whether applying the CAI or the CAI (over-intervention) with excessive intervention (hyper-parameters set to 2 times as the normal), the PPL, Coherence and Fluency scores remains stable without significant fluctuations. This indicates that the CAI method does not sacrifice the model's semantic coherence and contextual fluency.

\section{Domain Generalization Performance}
\label{sec:F}

\begin{table}[!ht]
\small
\centering
\renewcommand{\arraystretch}{1.0} 
\setlength{\tabcolsep}{6pt} 
\begin{tabular}{lccc}
\toprule
Domain & Dataset & Method & Accuracy  \\
\midrule
Medical & VQA-RAD & Greedy & 54.18\% \\
&&CAI & 58.17\%\\
\midrule
OCR & MMBench& Greedy & 74.31\%\\
&& CAI & 77.54\% \\
\bottomrule
\end{tabular}
\caption{Results on VQA-RAD and MMbench OCR subset.}. 
\label{domain}
\end{table}

In domain-specific tasks, the CAI method demonstrates certain generalization ability to some extent. Although caption queries are general instructions, they are extensively used during model pretraining. Activating the relevant attention patterns facilitates fine-grained visual information capture, thereby enhancing downstream task performance. To evaluate CAI's effectiveness in specific domains, we selected VQA-RAD \cite{lau2018dataset} from the medical domain and the MMBench \cite{liu2024mmbench} OCR subset. The experimental results of LLaVA-1.5-7b, as presented in the table \ref{domain}, show consistent improvements over the baseline, indicating the CAI method's generalization ability.

\section{Detailed Experimental Setup of Quantitative Analysis}
\label{sec:G}
We sample 1,000 images from the MS-COCO dataset \cite{lin2014microsoft}. For each image, we propose one caption query and two different non-caption queries (non-caption-1 \& non-caption-2) to analyze differences attributable to query types. 

We consider a LVLM parametrized by \( \theta \). The model receives as input a textual query $ \boldsymbol{T} = \{t_1, t_2, \dots, t_n\} $ and a visual input $ \boldsymbol{V} = \{v_1, v_2, \dots, v_m\} $, where $ n $ and $ m $ denote the sequence lengths of the text and visual inputs. The text and vision inputs are concatenated together to form the first layer input $ \boldsymbol{H}^{1} = \mathrm{concat}(\boldsymbol{V},\boldsymbol{T}) \in \mathbb{R}^{(m+n)\times d}$ for the $L$ layers $\times$ $H$ heads decoder. For an image, the last input token's visual attention weight of $H$-th head in $L$-th layer $\boldsymbol{Sum}_{(l,h)}$ can be computed as:  

\begin{equation}
\boldsymbol{A}_{(l,h)} = \textrm{softmax}( \frac{\boldsymbol{Q}_{(l,h)} \boldsymbol{K}_{(l,h)}^T}{\sqrt{d}}),
\end{equation}

\begin{equation}
\boldsymbol{Sum}_{(l,h)}=\sum_{i=1}^{m}\boldsymbol{A}^{-1}_{(l,h)}[i],
\end{equation}

where the $\boldsymbol{Q}_{(l,h)}$ and $\boldsymbol{K}_{(l,h)}$ are the Query and Key matrixs of the $k$-th head in $l$-th layer, $\boldsymbol{A}^{-1}_{(l,h)}[i]$ is the last input token's attention weight of the $i$-th input token. For a dataset of $B$ samples, the sum of visual attention weight can be computed as:

\begin{equation}
\boldsymbol{S}_{(l,h)}=\sum_{b=1}^{B}\boldsymbol{Sum}_{(l,h)}.
\end{equation}

Then we record the sum of visual attention weights from the last input token for three types of queries: $S^{cap}_{(l,h)}$ for caption query, $S^{non-1}_{(l,h)}$ for non-caption query 1 and $S^{non-2}_{(l,h)}$ for non-caption query 2. The head-wise Change Rate $Rate_{(l,h)}$ and layer-wise Change Rate $Rate_{(l)}$ can be computed as:

\begin{equation}
Rate_{(l,h)}=\frac{S^{cap}_{(l,h)} - S^{non-1}_{(l,h)}}{S^{non-1}_{(l,h)}},
\end{equation}

\begin{equation}
    Rate_{(l)}=\frac{\sum_{h=1}^{H}{(S^{cap}_{(l,h)} - S^{non-1}_{(l,h)})}}{\sum_{h=1}^{H}S^{non-1}_{(l,h)}}.
\end{equation}

By comparison, we find that visual attention across particular attention heads was significantly enhanced when fed caption compared to non-caption queries. These results provide strong support for our proposed motivation.

\section{Additional Case Studies}
\label{sec:H}
More case studies are shown as follows.

\begin{figure*}[!ht]
  \includegraphics[width=1.0\linewidth]{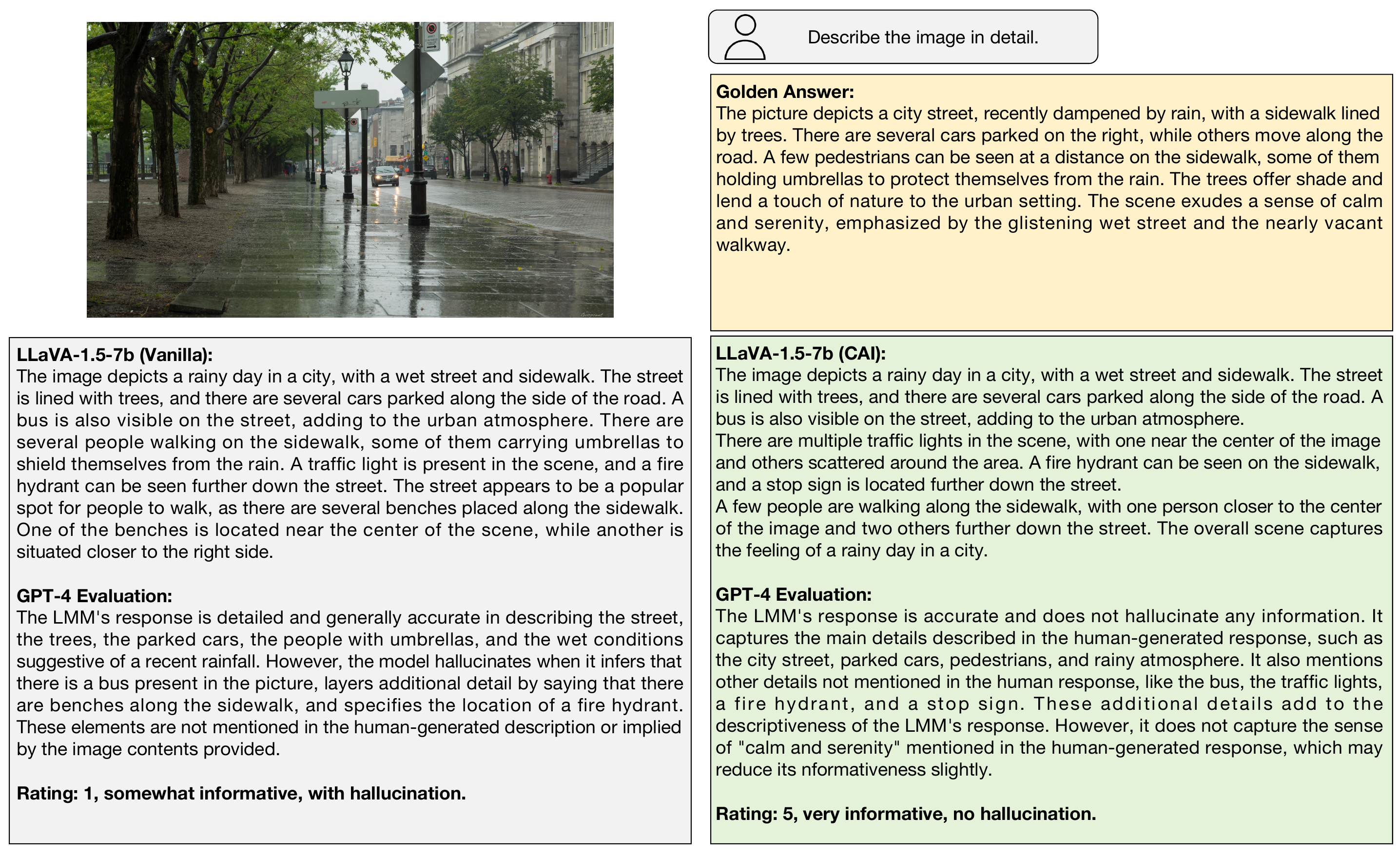}
  \caption{Caption query case of LLaVA-1.5-7b on MMHal-Bench.}
  \label{fig:case2}
\end{figure*}

\begin{figure*}[!ht]
  \includegraphics[width=1.0\linewidth]{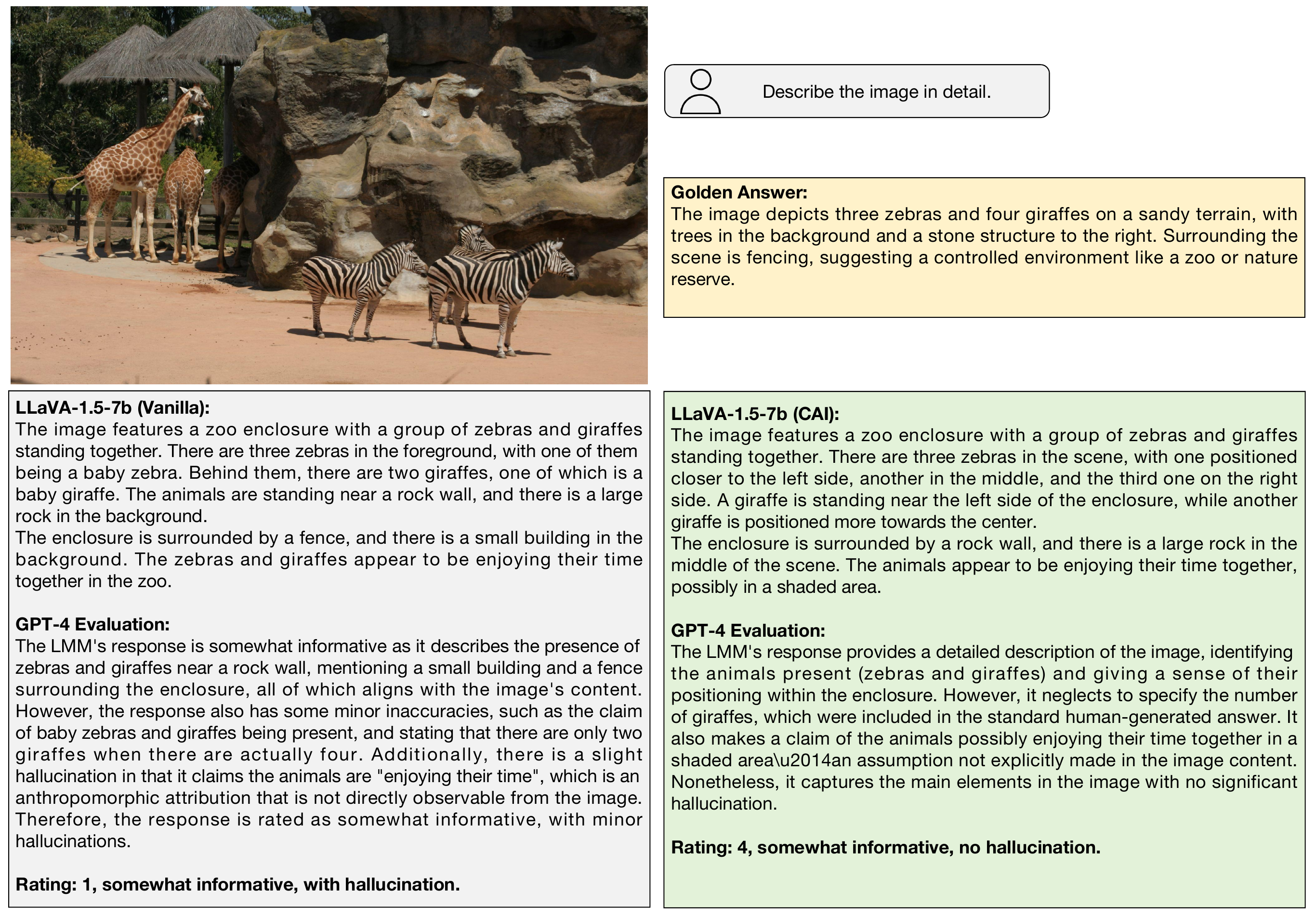}
  \caption{Another caption query case of LLaVA-1.5-7b on MMHal-Bench.}
  \label{fig:case3}
\end{figure*}

\begin{figure*}[!ht]
  \includegraphics[width=1.0\linewidth]{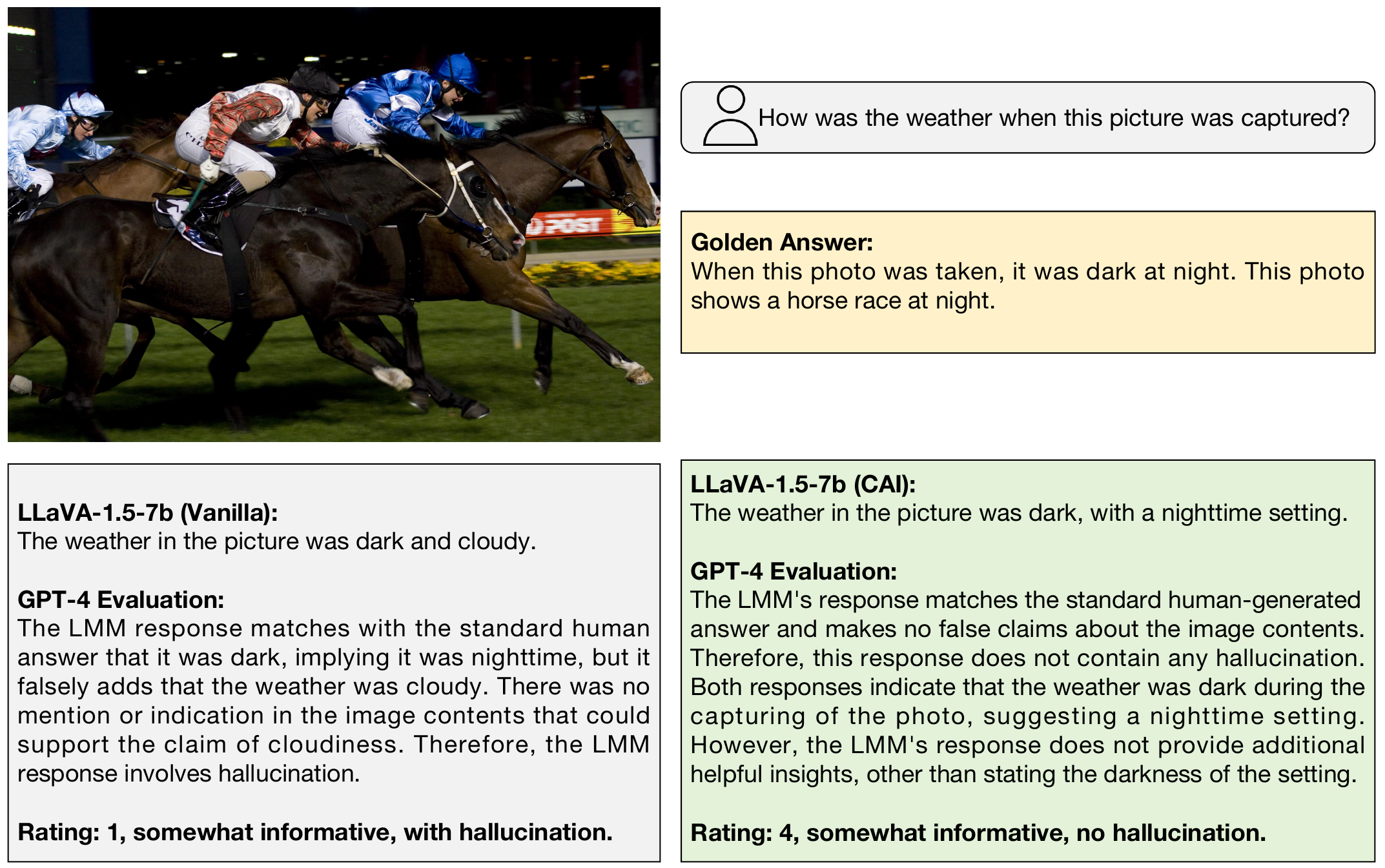}
  \caption{Non-caption query case of LLaVA-1.5-7b on MMHal-Bench.}
  \label{fig:case4}
\end{figure*}

\begin{figure*}[!ht]
  \includegraphics[width=1.0\linewidth]{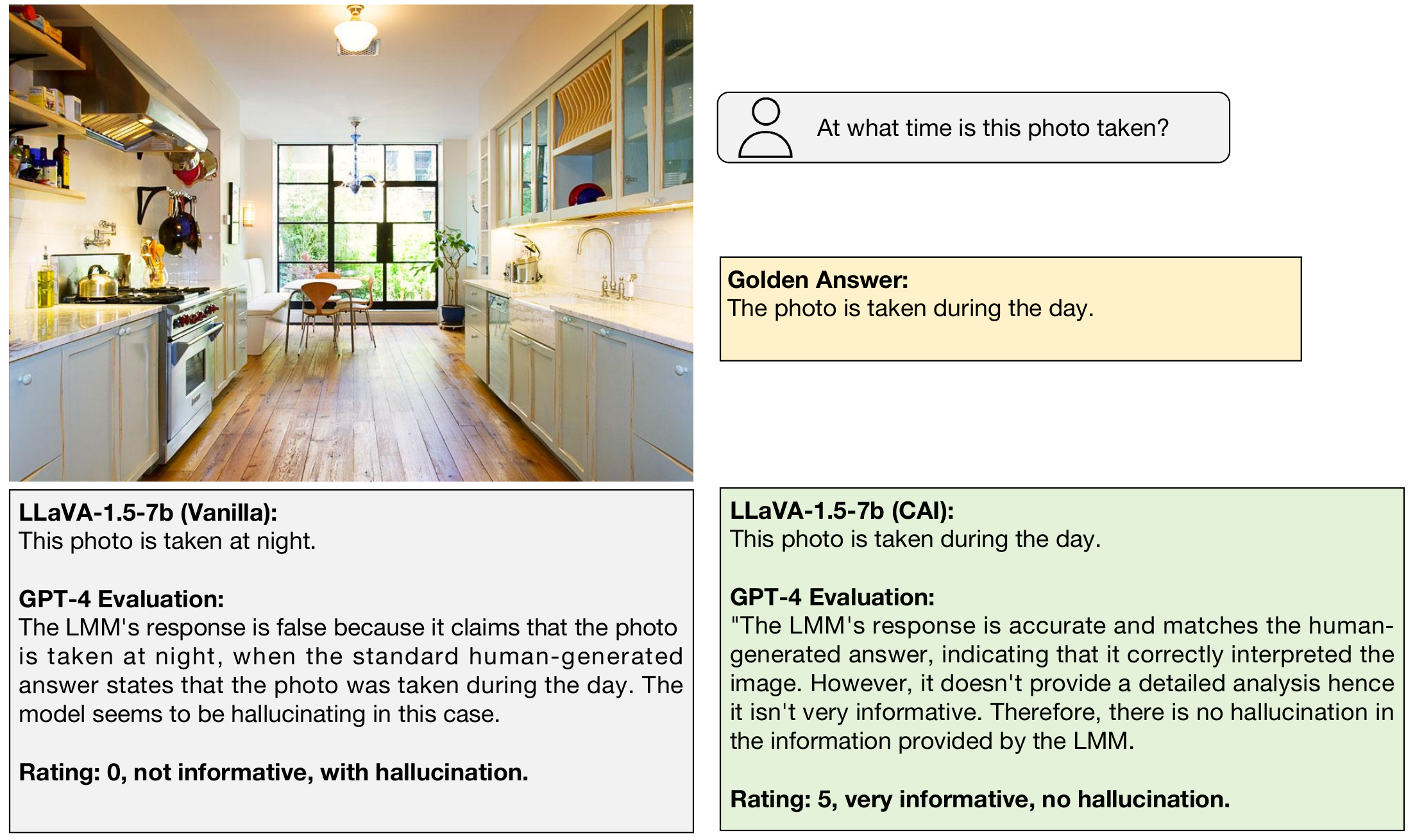}
  \caption{Non-caption query case of LLaVA-1.5-7b on MMHal-Bench.}
  \label{fig:case5}
\end{figure*}

\begin{figure*}[!ht]
  \includegraphics[width=1.0\linewidth]{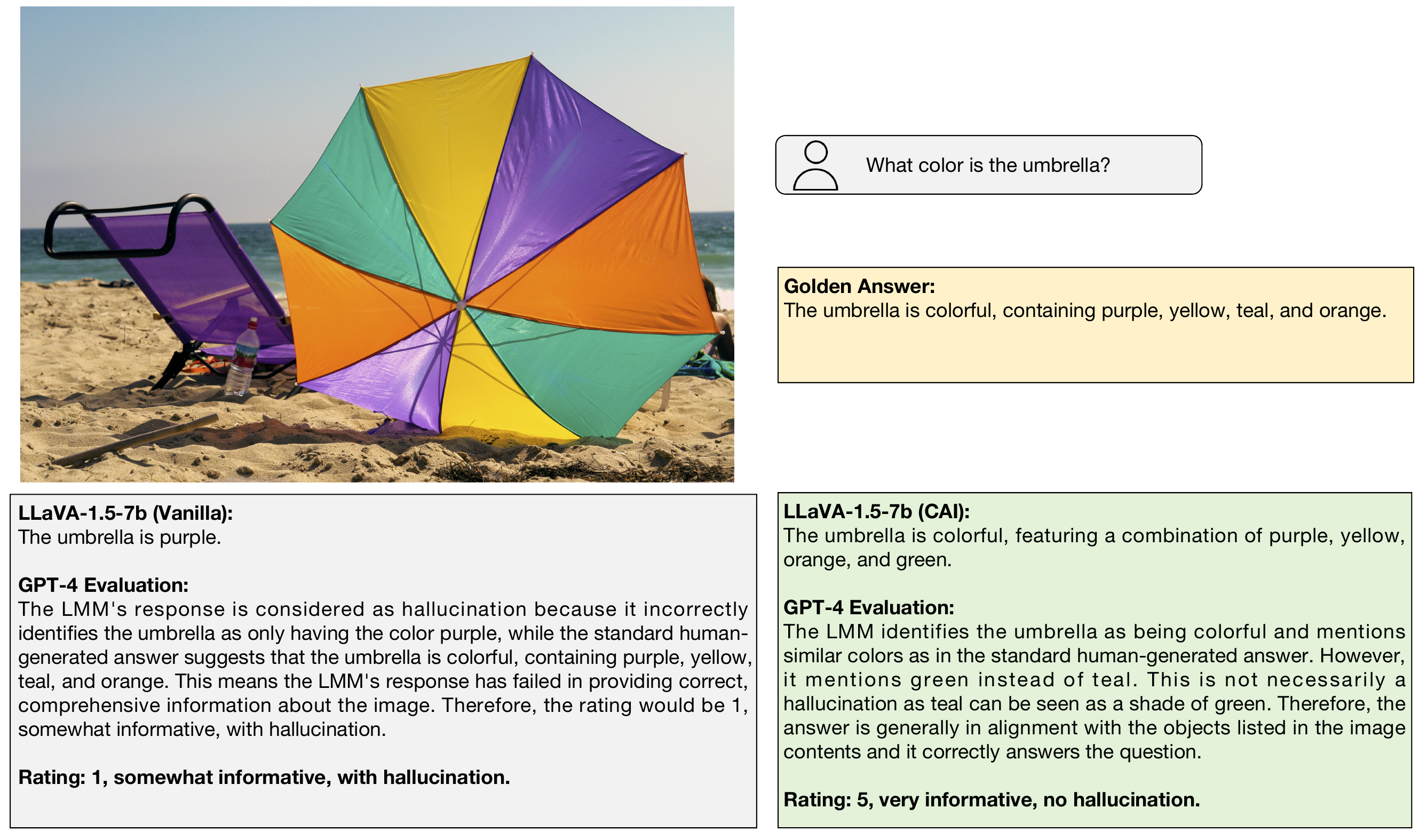}
  \caption{Non-caption query case of LLaVA-1.5-7b on MMHal-Bench.}
  \label{fig:case6}
\end{figure*}


\end{document}